\documentclass[conference]{IEEEtran}
\IEEEoverridecommandlockouts
\usepackage{cite}
\usepackage{amsmath,amssymb,amsfonts}
\usepackage{algorithmic}
\usepackage{graphicx}
\usepackage{textcomp}
\usepackage{xcolor}
\usepackage{times}
\usepackage{latexsym}
\usepackage{booktabs}
\usepackage{tablefootnote}
\usepackage[caption=false]{subfig}
\usepackage{amsthm}
\DeclareMathOperator*{\argmin}{arg\,min}
\usepackage[ruled, linesnumbered, vlined]{algorithm2e}
\SetAlFnt{\large}
\SetAlCapFnt{\small}
\SetAlCapNameFnt{\small}
\SetKwInput{KwInput}{Input}                
\SetKwInput{KwOutput}{Output}

\usepackage{url}

\def\BibTeX{{\rm B\kern-.05em{\sc i\kern-.025em b}\kern-.08em
    T\kern-.1667em\lower.7ex\hbox{E}\kern-.125emX}}
\begin{document}

\title{Learning Meta-Representations of One-shot Relations for Temporal Knowledge Graph Link Prediction}

\author{
    \IEEEauthorblockN{Zifeng Ding\IEEEauthorrefmark{1}\IEEEauthorrefmark{3}\IEEEauthorrefmark{2}, Bailan He\thanks{\IEEEauthorrefmark{1} Equal contribution.}\IEEEauthorrefmark{1}\IEEEauthorrefmark{3}\IEEEauthorrefmark{2}, Jingpei Wu\IEEEauthorrefmark{4}, Yunpu Ma\IEEEauthorrefmark{3}\IEEEauthorrefmark{2}, Zhen Han\thanks{\IEEEauthorrefmark{6} Corresponding author.}\IEEEauthorrefmark{6}\IEEEauthorrefmark{3}\IEEEauthorrefmark{2}, Volker Tresp\IEEEauthorrefmark{6}\IEEEauthorrefmark{3}\IEEEauthorrefmark{2}}
    \IEEEauthorblockA{\IEEEauthorrefmark{3}\textit{LMU Munich}}
    \IEEEauthorblockA{\IEEEauthorrefmark{2}\textit{Siemens AG}}
    \IEEEauthorblockA{\IEEEauthorrefmark{4}\textit{Technical University of Munich}
    \\\{zifeng.ding, bailan.he\}@campus.lmu.de, jingpei.wu@tum.de, cognitive.yunpu@gmail.com,
    \\ hanzhen02111@hotmail.com, volker.tresp@siemens.com}
}

\maketitle

\begin{abstract}
Few-shot relational learning for static knowledge graphs (KGs) has drawn greater interest in recent years, while few-shot learning for temporal knowledge graphs (TKGs) has hardly been studied. Compared to KGs, TKGs contain rich temporal information, thus requiring temporal reasoning techniques for modeling. This poses a greater challenge in learning few-shot relations in the temporal context. In this paper, we follow the previous work that focuses on few-shot relational learning on static KGs and extend two fundamental TKG reasoning tasks, i.e., interpolated and extrapolated link prediction, to the one-shot setting. We propose four new large-scale benchmark datasets and develop a TKG reasoning model for learning one-shot relations in TKGs. Experimental results show that our model can achieve superior performance on all datasets in both TKG link prediction tasks.
\end{abstract}


\section{Introduction}
Knowledge graphs (KGs) represent factual information in the form of triplets $(s,r,o)$, e.g., (\textit{Joe Biden}, \textit{is president of}, \textit{USA}), where $s$, $o$ are the subject and the object of a fact, and $r$ is the relation between $s$ and $o$. KGs have been extensively used to aid the downstream tasks in the field of artificial intelligence, e.g., recommender systems \cite{wang2019explainable} and question answering \cite{zhang2018variational,ding2022forecasting}. By incorporating time information into KGs, temporal knowledge graphs (TKGs) represent every fact with a quadruple $(s,r,o,t)$, where $t$ denotes the timestamp specifying the time validity of the fact. With the introduction of temporal constraints, TKGs are able to describe the ever-changing knowledge of the world. For example, due to the evolution of world knowledge, the fact (\textit{Angela Merkel}, \textit{is chancellor of}, \textit{Germany}) is valid only before (\textit{Olaf Scholz}, \textit{is chancellor of}, \textit{Germany}). TKGs naturally capture the evolution of relational facts in a time-varying context.


Though KGs are constructed with large-scale data, they still suffer from the problem of incompleteness \cite{min2013distant}. Hence, there has been extensive work aiming to propose KG reasoning models to infer the missing facts, i.e., the missing links, in KGs.
Similar to KGs, TKGs are also known to be highly incomplete. This draws huge attention to developing TKG reasoning methods for link prediction (LP) on TKGs. 
While a lot of proposed methods focus on predicting the links in the interpolation setting \cite{leblay2018deriving,lacroixtensor,wu2020temp,jung2021learning,ding2022simple}, where they predict missing facts at the observed timestamps, another line of work pays attention to forecasting
the TKG facts at the unobserved future timestamps and achieving extrapolation \cite{jin2020recurrent,zhu2021learning,han2021learning,han2021explainable}.
Most of these methods require a huge amount of data associated with each relation to learn expressive relation representations, however, it has been found that a large portion of KG and TKG relations are sparse (i.e., these relations are long-tail and only occur for a handful of times) \cite{xiong2018one,mirtaheri2021oneshot}.
This leads to the degenerated link inference performance of the traditional KG and TKG reasoning methods when they are predicting the links concerning sparse relations. To tackle this problem, a number of few-shot learning (FSL) methods \cite{xiong2018one,chen2019meta,zhang2020few,sheng2020adaptive} employ a meta-learning framework and learn to predict the unseen links concerning a sparse relation, given only very few observed KG facts associated to this relation. 
Based on these methods, \cite{mirtaheri2021oneshot} develops a method aiming to alleviate this problem for TKGs.
It formulates the one-shot TKG extrapolated LP task for learning the sparse relations in TKGs and proposes new datasets for it.

While \cite{mirtaheri2021oneshot} manages to generalize the extrapolated LP task to the one-shot setting, it still has limitations: (1) \cite{mirtaheri2021oneshot} fails to formulate the one-shot TKG interpolated LP task for the sparse relations. Since both interpolated and extrapolated LP serve as the fundamental tasks in TKG reasoning, it is also important to deal with the sparse relations in the interpolation setting; 
(2) For each to-be-predicted link $(s,r,o,t)$, traditional KG and TKG LP methods, e.g.,  \cite{bordes2013translating,leblay2018deriving}, consider predicting both the subject entity and the object entity, by deriving two LP queries $(s,r,?,t)$ and $(?,r,o,t)$. 
However, the previously-proposed KG and TKG few-shot relational learning methods, e.g.,  \cite{xiong2018one,mirtaheri2021oneshot}, only consider predicting the missing object entity, which makes the task settings unreasonable since both subject and object entity prediction are of great concern in KG and TKG LP; 
(3) The datasets proposed in \cite{mirtaheri2021oneshot} for the one-shot TKG extrapolated LP task have certain flaws. The number of the associated quadruples for each sparse relation is extremely small, which leads to incomprehensive training and evaluation data. Training with incomprehensive training sets would cause instability during training, and evaluating on the tiny evaluation sets makes it hard to determine the model performance accurately. Inaccurate validation results would be misleading in parameter optimization, and the models cannot be fairly judged with inaccurate test results.

To this end, we extend both TKG interpolated and extrapolated LP to the one-shot setting, and propose a model learning \textbf{m}eta-representations of \textbf{o}ne-\textbf{s}ho\textbf{t} relations for solving both tasks in TKGs (MOST). 
MOST learns the meta-representation of each sparse relation $r$ based on its associated one-shot TKG fact. It further employs a metric function to compute the plausibility scores of the unobserved facts concerning $r$.
The main contribution of our work (with corresponding sections) is summarized as follows: (1) We propose the one-shot TKG interpolated LP task. To the best of our knowledge, this is the first work generalizing TKG interpolated LP to the one-shot setting for predicting the links concerning sparse relations (Section \ref{sec: one shot setup});
(2) We fix the unreasonable task setting employed by the previous TKG one-shot relational learning method, and redefine the one-shot TKG extrapolated LP task. We conduct both subject and object entity prediction on the quadruples regarding sparse relations (Section \ref{sec: one shot setup});
(3) We propose four new large-scale datasets for one-shot relational learning on TKGs. For every sparse relation, we have a substantial number of associated TKG facts, which promotes reliable model training and evaluation (Section \ref{sec: new datasets});
(4) We propose a model solving both interpolated and extrapolated LP for one-shot relations on TKGs (Section \ref{sec: our method}). We evaluate our model on all four newly-proposed datasets and compare it with recent baselines. Our model achieves state-of-the-art performance on all datasets in both tasks (Section \ref{sec: experiments}).

\section{Preliminaries and Related Work}
\subsection{Temporal Knowledge Graph Reasoning} 
Let $\mathcal{E}$, $\mathcal{R}$ and $\mathcal{T}$ represent a finite set of entities, relations and timestamps, respectively. A temporal knowledge graph (TKG) $\mathcal{G}$ is a relational graph consisting of a finite set of facts denoted with quadruples in the form of $(s, r, o, t)$, i.e., $\mathcal{G} = \{(s,r,o,t)|s, o\in \mathcal{E}, r \in \mathcal{R}, t \in \mathcal{T}\}  \subseteq \mathcal{E}\times \mathcal{R} \times \mathcal{E} \times \mathcal{T}$.
A complete TKG $\mathcal{G}$ contains both the observed facts $\mathcal{G}_{obs}$ and the unobserved true facts $\mathcal{G}_{un}$, i.e., $\mathcal{G} = (\mathcal{G}_{obs} \cup \mathcal{G}_{un})$, where $\mathcal{G}_{obs} \cap \mathcal{G}_{un} = \emptyset$. Given $\mathcal{G}_{obs}$, TKG LP aims to predict the ground truth object (or subject) entities of LP queries $(s_q, r_q, ?, t_q)$ (or $(?, r_q, o_q, t_q)$), where $(s_q, r_q, o_q, t_q)\in \mathcal{G}_{un}$. The prediction can be based on all the observed facts $\{(s,r,o,t_i)|t_i\in \mathcal{T}\} \subseteq \mathcal{G}_{obs}$ from any timestamp in interpolated LP, while extrapolated LP regulates that the prediction can only be based on the observed facts $\{(s,r,o,t_i)|t_i<t_q\} \subseteq \mathcal{G}_{obs}$ appearing before the query timestamp $t_q$. Both of two TKG LP tasks are fundamental in TKG reasoning. Recently, extensive studies have been done for both interpolated LP \cite{leblay2018deriving,lacroixtensor,wu2020temp,jung2021learning,ding2022simple} and extrapolated LP \cite{jin2020recurrent,zhu2021learning,han2021learning,han2021explainable}.
In these researches, TKG interpolated LP is also termed as TKG completion and TKG extrapolated LP is also termed as TKG forecasting or TKG link forecasting.

\subsection{Few-Shot Relational Learning for Knowledge Graphs}
Few-shot learning (FSL) is a type of machine learning problems where models are asked to perform well on the unobserved data examples for each class, given only a few labeled class-specific observed data examples. When the number of the labeled examples equals 1, FSL problems become one-shot learning problems. 
Meta-learning approaches aim to quickly learn novel concepts (with only a few concept-related data examples) by generalizing from previously encountered learning tasks \cite{laenen2021episodes}, which fits well to solving the problems in the few-shot setting. 
Episodic training \cite{vinyals2016matching} is a meta-learning framework, where a model is trained over episodes. Each episode can be considered as a mini-training process on a training task $T$, where a number of "training examples" (support set $\mathcal{S}$) and "test examples" (query set $\mathcal{Q}$) are sampled and a loss function $l_{\theta}$ is calculated over $\mathcal{Q}$ conditioned on $\mathcal{S}$. $\theta$ denotes the model parameters. With episodic training, a model is trained over a large set of training tasks to explicitly learn to learn from a given support set to minimise a loss over a batch of examples in the query set. Assume we have a large set of training tasks $\mathbb{T} = \{T_i\}_{i=1}^N$, where $T_i = \{\mathcal{S}_i, \mathcal{Q}_i\}$ and $N$ is the total number of the training tasks, the training objective of a model is given as $\theta = \argmin_{\theta}\mathbb{E}_{T_i\sim \mathbb{T}} \left[ \frac{1}{|\mathcal{Q}_i|}\sum_{q \in \mathcal{Q}_i}
    \left[ l_{\theta}(q | \mathcal{S}_i)\right] \right]$. 
$q$ denotes a data example in the query set $\mathcal{Q}_i$. Episodic training manages to simulate the few-shot situation when only a small number of data examples are sampled to form the support set of each training task $T$, 
thus serving as a common meta-learning paradigm to solve FSL problems.

To better learn the sparse relations in KGs, \cite{xiong2018one} first introduces an FSL problem, i.e., few-shot KG LP, where models are asked to infer unobserved KG facts for each sparse relation $r$ conditioned on only a few observed KG facts concerning $r$. It further formulates few-shot KG LP into a meta-learning problem and trains its proposed method Gmatching using episodic training. Several researches follow \cite{xiong2018one} and employ episodic training to train different FSL models for solving the few-shot KG LP task \cite{chen2019meta,zhang2020few,sheng2020adaptive,niu2021relational}. 
\cite{mirtaheri2021oneshot} first introduces one-shot relational learning into TKGs and proposes one-shot TKG extrapolated LP. Following \cite{xiong2018one}, it formulates the one-shot TKG extrapolated LP task into a meta-learning problem and employs episodic training for training its model OAT. 
Among these methods, GMatching \cite{xiong2018one}, FSRL \cite{zhang2020few}, FAAN \cite{sheng2020adaptive} and OAT \cite{mirtaheri2021oneshot} are metric-based meta-learning approaches that use metric functions to do similarity matching of the few-shot examples and the to-be-predicted links. 
MetaR \cite{chen2019meta} and GANA \cite{niu2021relational} are optimization-based meta-learning approaches that employ Model-Agnostic Meta-Learning \cite{finn2017model} for FSL. They learn a good initialization of model parameters and achieve fast adaption of them with very few relation-specific examples. 

Recently, another line of work aims at learning newly-emerged few-shot entities in TKGs. \cite{ding2022few} proposes an FSL problem, i.e., TKG few-shot out-of-graph (OOG) LP, that generalizes TKG interpolated LP to the few-shot setting. A model called FILT is trained with episodic training to solve the problem. To improve performance on TKG few-shot OOG LP, \cite{ding2023improving} designs a meta-learning-based model using confidence-augmented reinforcement learning. \cite{wanglearning} proposes another FSL problem, i.e., few-shot TKG reasoning, that extends the extrapolation LP setting to an FSL problem. Wang et al. develop MetaTKGR that addresses both few-shot and time shift challenges. These methods all focus on few-shot unseen entities, but cannot deal with sparse relations.
\section{One-Shot Temporal Knowledge Graph Link Prediction Setup}
\label{sec: one shot setup}
\begin{figure*} 
    \centering
  \subfloat[Example of one-shot TKG LP tasks.\label{fig: example of oneshot lp}]{%
       \includegraphics[width=0.68\textwidth]{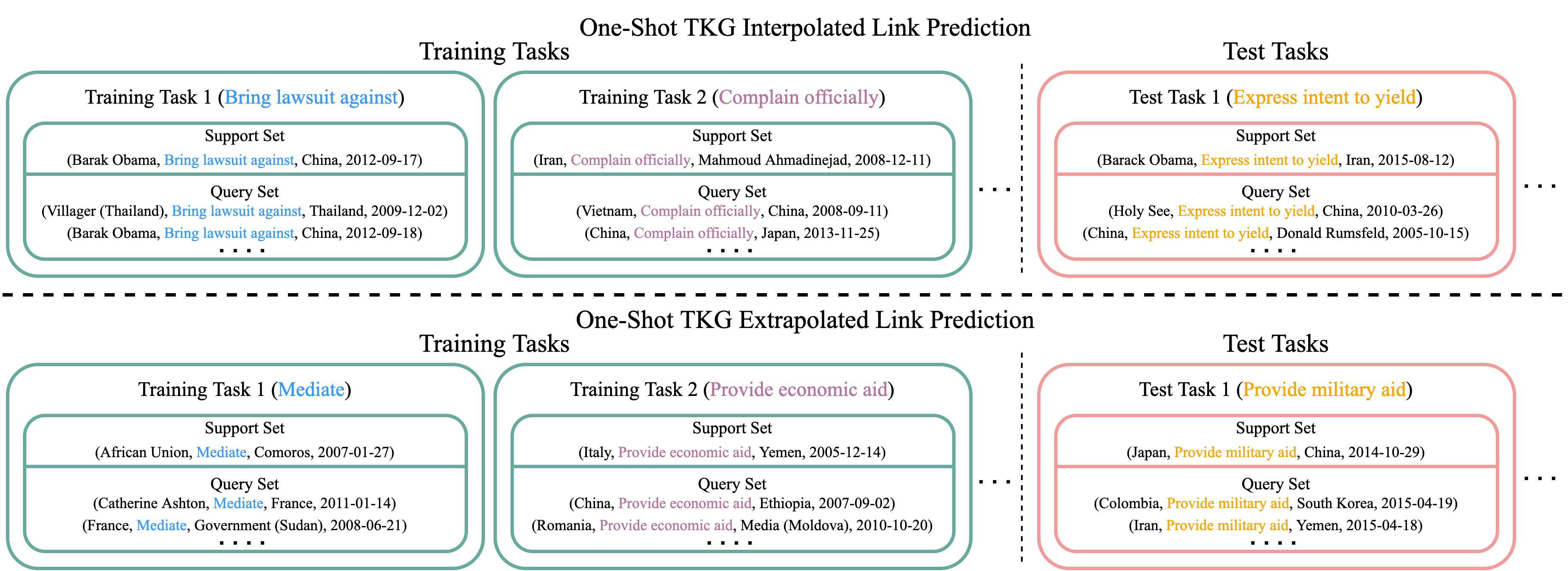}}
    \hfill
  \subfloat[Quadruple timestamps of meta-learning sets in both tasks.\label{fig: time overlap}]{%
        \includegraphics[width=0.30\textwidth]{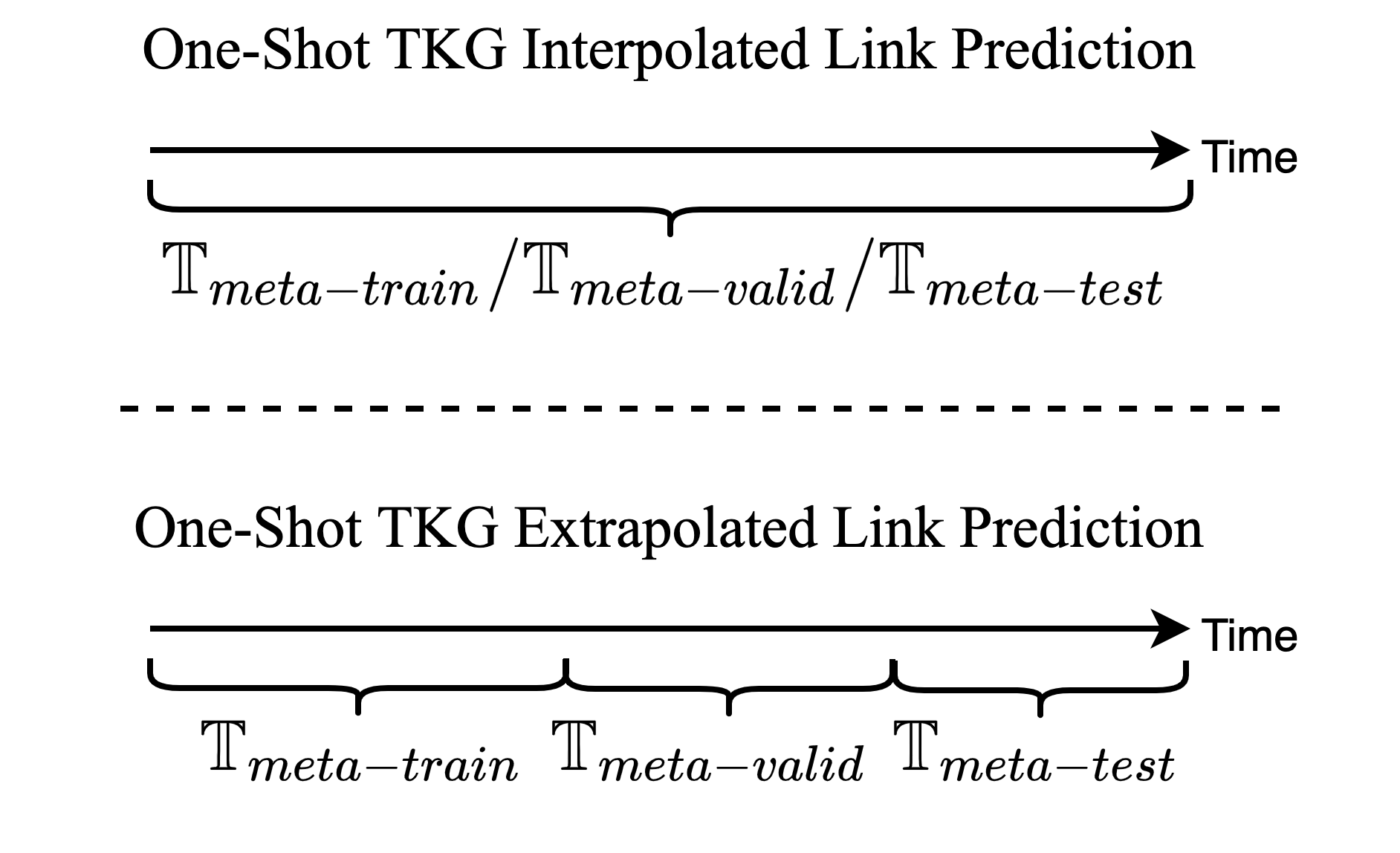}}
    \hfill
  \caption{(a) Example of one-shot TKG LP tasks. Models are trained over training tasks using episodic training and then evaluated with validation and test tasks (validation tasks are omitted in the figure for brevity). Each training (or validation, test) task $T_r$ corresponds to a sparse relation $r$. In the one-shot extrapolation setting, $T_r$'s support timestamp $t_0 < \min(\{t_q|(s_q,r,o_q,t_q) \in \mathcal{Q}_r\})$. (b) Meta-learning sets share the same time span in interpolation, while they are arranged in a non-overlapped sequential order in extrapolation. 
  For one-shot extrapolation, the maximum timestamp of the quadruples in $\mathbb{T}_{meta-train}$ is smaller than the minimum timestamp of the quadruples in $\mathbb{T}_{meta-valid}$, and the maximum timestamp of the quadruples in $\mathbb{T}_{meta-valid}$ is smaller than the minimum timestamp of the quadruples in $\mathbb{T}_{meta-test}$.}
  \label{fig: inter vs extra} 
\end{figure*}
We give the definition of our newly-proposed task one-shot TKG interpolated LP, and redefine one-shot TKG extrapolated LP to consider both subject and object entity prediction.
For a TKG $\mathcal{G}$, all its relations $\mathcal{R}$ can be classified into two groups, i.e., frequent relations $\mathcal{R}_{freq}$ and sparse relations $\mathcal{R}_{sp}$, where $\mathcal{R}_{freq} \cap \mathcal{R}_{sp} = \emptyset$ and $\mathcal{R} = (\mathcal{R}_{freq} \cup \mathcal{R}_{sp})$. A background graph $\mathcal{G}' \subseteq \mathcal{E}\times \mathcal{R}_{freq} \times \mathcal{E} \times \mathcal{T}$ is constructed by including all the quadruples concerning frequent relations, where $\mathcal{G}' \subseteq \mathcal{G}$.

\noindent
\textbf{Definition 1 (One-Shot TKG Interpolated Link Prediction).} 
Assume we observe only one quadruple $(s_0,r,o_0,t_0)$ corresponding to each sparse relation $r$, where $r \in \mathcal{R}_{sp}$, $s_0, o_0 \in \mathcal{E}$ and $t_0 \in \mathcal{T}$. Given $(s_0,r,o_0,t_0)$ and the whole background graph $\mathcal{G}'$, one-shot TKG interpolated LP aims to predict the missing entity of each LP query, i.e., $(s_q,r,?,t_q)$ or $(?,r,o_q,t_q)$, derived from the unobserved quadruples containing $r$, where $s_q, o_q \in \mathcal{E}$ and $t_q \in \mathcal{T}$.

\noindent
\textbf{Definition 2 (One-Shot TKG Extrapolated Link Prediction).} 
Assume we observe only one quadruple $(s_0,r,o_0,t_0)$ corresponding to each sparse relation $r$, where $r \in \mathcal{R}_{sp}$, $s_0, o_0 \in \mathcal{E}$ and $t_0 \in \mathcal{T}$. Given $(s_0,r,o_0,t_0)$, together with a set of observed TKG facts that appear prior to $t_0$ and belong to the background graph $\mathcal{G}'$, one-shot TKG extrapolated LP aims to predict the missing entity of each LP query, i.e., $(s_q,r,?,t_q)$ or $(?,r,o_q,t_q)$, derived from the unobserved quadruples containing $r$, where $s_q, o_q \in \mathcal{E}$, $t_q \in \mathcal{T}$ and $t_q > t_0$.

In our definitions, we consider both subject and object prediction, which fixes the unreasonable setting of previous few-shot relational learning methods that neglect subject prediction. To achieve subject prediction, following previous TKG reasoning methods, e.g., \cite{han2021learning}, we add reciprocal relations for every quadruple, i.e., adding $(o,r^{-1},s,t)$ for every $(s,r,o,t)$. If $r \in \mathcal{R}_{sp}$, we treat $r^{-1}$ as a separate sparse relation and transform every subject prediction query $(?,r,o,t)$ of $r$ to an object prediction query $(o,r^{-1},?,t)$ of $r^{-1}$.
$r$ and $r^{-1}$ are in the same split, e.g., if $r$ belongs to sparse training relations $\mathcal{R}_{sp}^{train}$, then $r^{-1} \in \mathcal{R}_{sp}^{train}$.

We further formulate the one-shot TKG LP tasks into meta-learning problems. Following  \cite{xiong2018one,mirtaheri2021oneshot}, we assume that we have access to a set of training tasks for episodic training. Each training task $T_r$ corresponds to a sparse relation $r\in \mathcal{R}_{sp}^{train}$ ($\mathcal{R}_{sp}^{train} \subset \mathcal{R}_{sp}$). $T_r = \{\mathcal{S}_r, \mathcal{Q}_r\}$, where $\mathcal{S}_r$ is the support set of $T_r$ containing only one support quadruple $(s_0,r,o_0,t_0)$, and $\mathcal{Q}_r = \{(s_q, r, o_q, t_q)\}$ is the query set of $T_r$ containing a number of $r$-related quadruples other than $(s_0,r,o_0,t_0)$. The set of all training tasks is denoted as the meta-training set $\mathbb{T}_{meta-train}$. A loss function $l_\theta((s_q, r, o_q, t_q)|\mathcal{S}_r)$ is used to indicate how well the TKG reasoning model works on the query quadruple $(s_q, r, o_q, t_q)$, given the support set $\mathcal{S}_r$. $\theta$ denotes the model parameters. The training objective of the model is given as $\theta = \argmin_{\theta} \mathbb{E}_{T_r\sim \mathbb{T}_{meta-train}} \left[ \frac{1}{|\mathcal{Q}_r|}\sum_{q \in \mathcal{Q}_r}
    \left[ l_{\theta}(q | \mathcal{S}_r)\right] \right]$, 
where $q$ represents a query quadruple $(s_q,r,o_q,t_q)$. $T_r$ is sampled from the meta-training set $\mathbb{T}_{meta-train}$, and $|\mathcal{Q}_r|$ denotes the number of the query quadruples regarding the sparse relation $r$.
After training, the TKG reasoning model will be evaluated on a meta-test set $\mathbb{T}_{meta-test}$ corresponding to unseen sparse relations $\mathcal{R}_{sp}^{test}$, where $\mathcal{R}_{sp}^{test} \subset \mathcal{R}_{sp}$ and $\mathcal{R}_{sp}^{train} \cap \mathcal{R}_{sp}^{test} = \emptyset$. We also validate the model performance with a meta-validation set $\mathbb{T}_{meta-valid}$ ($\mathcal{R}_{sp}^{valid} \subset \mathcal{R}_{sp}$, $\mathcal{R}_{sp}^{train} \cap \mathcal{R}_{sp}^{valid} = \emptyset, \mathcal{R}_{sp}^{valid} \cap \mathcal{R}_{sp}^{test} = \emptyset$). Similar to meta-training, for each sparse relation in meta-validation and meta-test, only one associated quadruple serves as its support set, and all the links in its query set are to be predicted.
For each sparse relation $r$, in the interpolated LP, there is no constraint for the support timestamp $t_0$, while in the extrapolated LP, temporal constraint is imposed that $t_0 < \min(\{t_q|(s_q,r,o_q,t_q) \in \mathcal{Q}_r\})$. We present one example in Fig. \ref{fig: example of oneshot lp} for each of the one-shot TKG LP task. Following \cite{mirtaheri2021oneshot}, to prevent exposing the models to the future information, in the extrapolation task, we further keep the spans of the quadruples' timestamps from different meta-learning sets ($\mathbb{T}_{meta-train}$, $\mathbb{T}_{meta-valid}$, $\mathbb{T}_{meta-test}$) in a non-overlapped sequential order (Fig. \ref{fig: time overlap}).
\section{New Datasets and Our Method}
\label{sec: new datasets and our method}
\subsection{Proposing New Datasets}
\label{sec: new datasets}
\begin{figure} 
    \centering
  \subfloat[Histogram of sparse relation frequency on ICEWS-based datasets.\label{fig: icews_comparison}]{%
       \includegraphics[width=0.88\columnwidth]{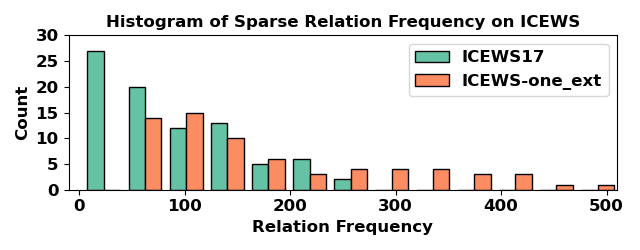}}
    \hfill
    \\
  \subfloat[Histogram of sparse relation frequency on GDELT-based datasets.\label{fig: gdelt_comparison}]{%
        \includegraphics[width=0.88\columnwidth]{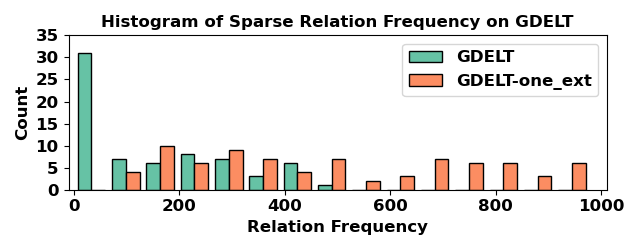}}
    \hfill
  \caption{Sparse Relation frequency comparison between ICEWS-one\_ext and ICEWS17; GDELT-one\_ext and GDELT.}
  \label{fig: data comparison} 
\end{figure}

By taking subsets of two benchmark TKG databases, i.e., ICEWS \cite{DVN/28075_2015} and GDELT \cite{leetaru2013gdelt}, Mirtaheri et al. \cite{mirtaheri2021oneshot} propose two one-shot extrapolated LP datasets, i.e., ICEWS17 and GDELT. They first set upper and lower thresholds, and then select the relations with frequency between them as sparse relations (frequency 50 to 500 for ICEWS17, 50 to 700 for GDELT). To prevent time overlaps among meta-learning sets (Fig. \ref{fig: time overlap}), they further remove a significant number of quadruples regarding sparse relations. Assume a relation $r$ is selected as a sparse relation and $T_r \in \mathbb{T}_{meta-train}$. The ending timestamp of the meta-training set is $t_1$. Then all the quadruples in $\{(s,r,o,t)|s,o \in \mathcal{E}, t> t_1\}$ are removed from the dataset. This leads to a considerably smaller query set $\mathcal{Q}_r$ when a large number of $r$-related facts take place after $t_1$. If $r$'s frequency is close to the lower threshold before removal, it is very likely that after removal, the number of associated quadruples left in $\{(s,r,o,t)|s,o \in \mathcal{E}, t\leq t_1\}$ becomes extremely small, leading to a tiny $\mathcal{Q}_r$ that causes instability during training. Similarly, if $T_r \in (\mathbb{T}_{meta-valid} \cup \mathbb{T}_{meta-test})$, evaluation over a tiny $\mathcal{Q}_r$ makes it hard to accurately determine the model performance since the test data is incomprehensive. Fig. \ref{fig: data comparison} shows that a large portion of sparse relations in ICEWS17 and GDELT have very few associated quadruples. To be specific, in ICEWS17, 31 out of 85 sparse relations have less than 50 associated quadruples, and in GDELT, 24 out of 69 sparse relations have less than 50 associated quadruples. Moreover, 4 out of 14 test relations have even less than 10 associated quadruples in ICEWS17, and this also applies for 11 out of 14 test relations in GDELT. This indicates that the datasets proposed in \cite{mirtaheri2021oneshot} have certain flaws.

To overcome these problems, we construct two new large-scale extrapolation LP datasets, i.e., ICEWS-one\_ext and GDELT-one\_ext, also by taking subsets from ICEWS and GDELT. ICEWS-one\_ext is constructed with the timestamped political facts happening from 2005 to 2015, while GDELT-one\_ext is constructed with the global social facts from Jan. 1, 2018 to Jan. 31, 2018. For sparse relation selection, we set the upper and lower thresholds of frequency to 100 and 1000 for ICEWS-one\_ext, 200 and 2000 for GDELT-one\_ext, and then split these relations into train/valid/test groups. We take the relations with higher frequency as frequent relations $\mathcal{R}_{freq}$ and build background graphs $\mathcal{G}'$ with all the quadruples containing them. Following \cite{mirtaheri2021oneshot}, we then remove a part of quadruples associated with sparse relations to prevent time overlaps among meta-learning sets. After removal, we further discard the relations with too few associated quadruples (less than 50 for ICEWS-one\_ext, 100 for GDELT-one\_ext). In this way, we prevent including meta-tasks $T_r$ with extremely small query set $\mathcal{Q}_r$.
From Fig. \ref{fig: data comparison}, we observe that ICEWS-one\_ext and GDELT-one\_ext have a substantial number of associated quadruples for each sparse relation, which promotes reliable model training and evaluation.
We also construct two datasets, i.e., ICEWS-one\_int and GDELT-one\_int, for one-shot TKG interpolated LP. Since in 
interpolation, models are allowed to be exposed to the information from any timestamp, we do not have to remove quadruples to eliminate time overlaps among meta-learning sets. To this end, we set the upper and lower thresholds of sparse relations' frequency to 50 and 500 for ICEWS-one\_int, 100 and 1000 for GDELT-one\_int, and then split these relations into train/valid/test groups. Statistics of our datasets are presented in Table \ref{tab: data}.
\begin{table}[htbp]
\caption{Dataset statistics (without reciprocal relations). $|\mathcal{R}_{sp}|$ denotes the number of sparse relations in $\mathcal{R}_{sp}^{train}$, $\mathcal{R}_{sp}^{valid}$, $\mathcal{R}_{sp}^{test}$.}
\label{tab: data}
    \centering
    \small\resizebox{0.8\columnwidth}{!}{
\begin{tabular}{c c c c c c c} \hline
Dataset&$|\mathcal{E}|$&$|\mathcal{R}|$&$|\mathcal{T}|$&$|\mathcal{R}_{sp}|$ & $|\mathcal{G}'|$\\ \hline
ICEWS-one\_ext  & $7,934$ & $109$ & $4,017$ & $53/6/11$ & 408,760\\ 
ICEWS-one\_int  & $10,356$ & $155$ & $4,017$ & $74/9/10$ & 441,553\\ 
GDELT-one\_ext  & $6,647$ & $155$ & $2,751$ & $55/7/11$  & 2,205,316\\ 
GDELT-one\_int  & $7,677$ & $181$ & $2,751$ & $64/8/8$ & 2,237,534\\
\hline
\end{tabular}}
\end{table}
\subsection{Our Method}
\label{sec: our method}
\begin{figure*}[htbp]
\centering
\includegraphics[scale=0.26]{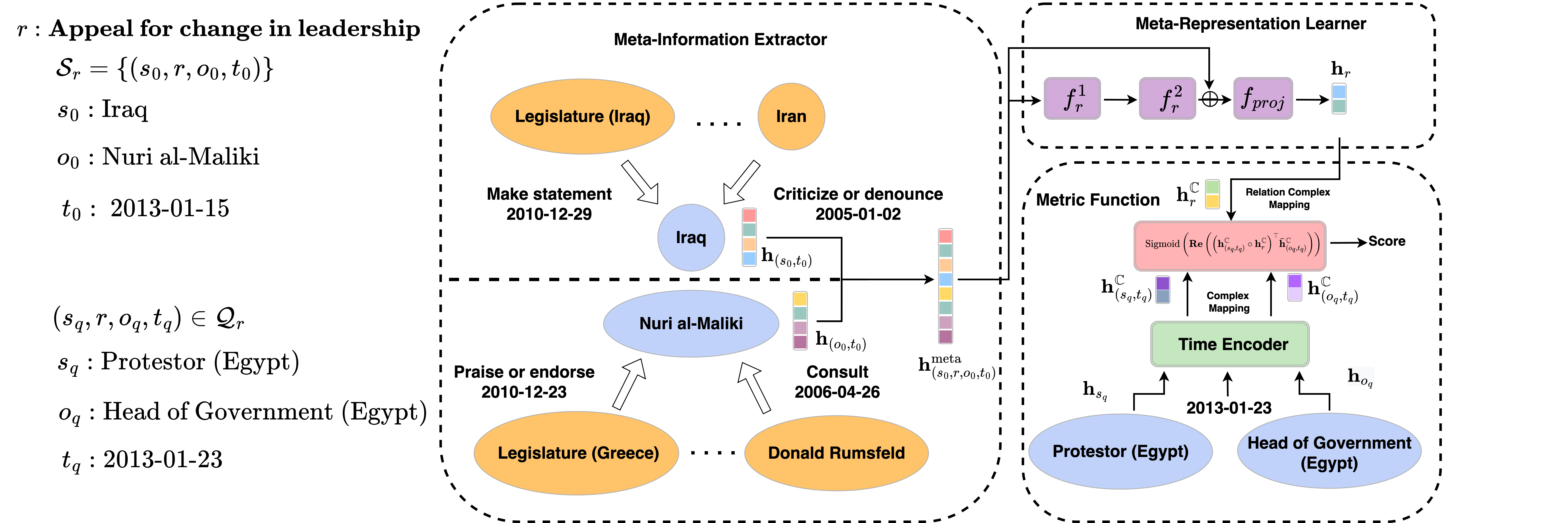}
\caption{\label{fig: model} Overview of MOST. It extracts meta-information from support entities, learns the meta-representation of the sparse relation and uses it to predict related links. Better viewed with the equations and notations in Section \ref{sec: our method}.} 
\end{figure*}
We propose a metric-based meta-learning model, i.e., MOST, to solve both one-shot TKG interpolated and extrapolated LP. 
Fig. \ref{fig: model} shows the overview of MOST. Given a support quadruple $(s_0,r,o_0,t_0)$ of the sparse relation $r$, MOST extracts the meta-information from the time-aware representations of support entities $s_0$, $o_0$ with a meta-information extractor, and uses a meta-representation learner to learn $r$'s meta-representation based on it. A metric function is then employed to compute the plausibility scores of TKG quadruples concerning $r$.
\subsubsection{Meta-Information Extractor}
Given the support quadruple $(s_0,r,o_0,t_0)$ of a sparse relation $r$, MOST aims to extract the meta-information of $r$ from its support entities $s_0$, $o_0$. MOST first employs a time-aware relational graph encoder to learn the contextualized time-aware entity representations of support entities. For every support entity ($s_0$ or $o_0$), MOST finds its temporal neighbors. It searches for the available background facts whose object entity corresponds to this support entity, and constructs a temporal neighborhood, e.g., $s_0$'s temporal neighborhood is denoted as $\mathcal{N}_{s_0} = \{(e',r', t')|r' \in \mathcal{R}_{freq}, (e', r', s_0, t') \in \mathcal{G}'\}$. 
It keeps a fixed number of temporal neighbors nearest to the support timestamp $t_0$ (MOST ensures the kept neighbors are prior to $t_0$ for one-shot TKG extrapolated LP). The number of sampled neighbors is a hyperparameter and can be tuned. We denote the filtered neighborhood as $\Tilde{\mathcal{N}}_{s_0}$ and $\Tilde{\mathcal{N}}_{o_0}$.
MOST then computes the time-aware representations of the support entities by aggregating the information provided by their temporal neighbors. The time-aware entity representations $\mathbf{h}_{(s_0,t_0)}$, $\mathbf{h}_{(o_0,t_0)}$ of $s_0$, $o_0$ are derived as follows: 
\begin{equation}
\label{eq: agg}
\resizebox{0.9\columnwidth}{!}{%
$
\begin{aligned}
    &\mathbf{h}_{(s_0,t_0)} = \mathbf{h}_{s_0}+ \delta_1 \sigma ( \frac{1}{|\Tilde{\mathcal{N}}_{s_0}|} \sum_{(e',r', t') \in \Tilde{\mathcal{N}}_{s_0}}\mathbf{W}_g \left(f(\mathbf{h}_{e'}\|\boldsymbol{\Phi}(t'))\circ\mathbf{h}_{r'}\right)),\\
    &\mathbf{h}_{(o_0,t_0)} = \mathbf{h}_{o_0}+ \delta_1 \sigma ( \frac{1}{|\Tilde{\mathcal{N}}_{o_0}|} \sum_{(e',r', t') \in \Tilde{\mathcal{N}}_{o_0}}\mathbf{W}_g \left(f(\mathbf{h}_{e'}\|\boldsymbol{\Phi}(t'))\circ\mathbf{h}_{r'}\right)).
\end{aligned}
$
}
\end{equation}
$\mathbf{h}_{s_0} \in \mathbb{R}^d$ and $\mathbf{h}_{o_0} \in \mathbb{R}^d$ denote the time-invariant entity representations of $s_0$ and $o_0$, respectively. $\mathbf{h}_{r'}\in \mathbb{R}^d$ denotes the relation representation of the frequent relation $r'$. $d$ is the dimension of the representations. Moreover, $\circ$, $\|$ represent Hadamard product and concatenation operation, respectively. $\mathbf{W}_g \in \mathbb{R}^{d\times d}$ is a weight matrix that processes the information in the graph aggregation. $f:\mathbb{R}^{2d} \to \mathbb{R}^{d}$ is a layer of feed-forward neural network. $\delta_1$ is a trainable parameter deciding how much information from the temporal neighbors is included in updating entity representations. $\sigma$ is an activation function. $\boldsymbol{\Phi}(t')$ denotes the time encoding function that encodes timestamp $t'$ as $\boldsymbol{\Phi}(t') = \sqrt{\frac{1}{d}}[cos(\omega_1t'+\phi_1), \dots, cos(\omega_{d}t'+ \phi_{d}))]$, 
where $\omega_1 \dots \omega_{d}$ and $\phi_1 \dots \phi_{d}$ are trainable parameters. We name our model with this timestamp encoder as MOST-TA. Besides, we develop another model variant MOST-TD by encoding time differences instead of timestamps. We input $t_0-t'$ instead of $t'$ into the time encoder. 
\begin{equation}
\label{eq: agg2}
\resizebox{0.95\columnwidth}{!}{%
$
\begin{aligned}
    &\mathbf{h}_{(s_0,t_0)} = \mathbf{h}_{s_0}+ \delta_1 \sigma ( \frac{1}{|\Tilde{\mathcal{N}}_{s_0}|} \sum_{(e',r', t') \in \Tilde{\mathcal{N}}_{s_0}}\mathbf{W}_g \left(f(\mathbf{h}_{e'}\|\boldsymbol{\Phi}(t_0 -t'))\circ\mathbf{h}_{r'}\right)),\\
    &\mathbf{h}_{(o_0,t_0)} = \mathbf{h}_{o_0}+ \delta_1 \sigma ( \frac{1}{|\Tilde{\mathcal{N}}_{o_0}|} \sum_{(e',r', t') \in \Tilde{\mathcal{N}}_{o_0}}\mathbf{W}_g \left(f(\mathbf{h}_{e'}\|\boldsymbol{\Phi}(t_0 -t'))\circ\mathbf{h}_{r'}\right)).
\end{aligned}
$
}
\end{equation}
We show in experiments (Section \ref{sec:exp results}) that both MOST-TA and MOST-TD can achieve state-of-the-art performance in one-shot TKG LP tasks.
After obtaining $\mathbf{h}_{(s_0,t_0)}$ and $\mathbf{h}_{(o_0,t_0)}$, we compute the meta-information of $r$ as
\begin{equation}
    \mathbf{h}^{\text{meta}}_{(s_0,r,o_0,t_0)} = \mathbf{h}_{(s_0,t_0)} \| \mathbf{h}_{(o_0,t_0)}.
\end{equation}
$\mathbf{h}^{\text{meta}}_{(s_0,r,o_0,t_0)} \in \mathbb{R}^{2d}$ represents the meta-information of $r$, given the support quadruple $(s_0,r,o_0,t_0)$.


\subsubsection{Meta-Representation Learner}
In the meta-representation learner, MOST derives the meta-representation of $r$ given the meta-information $\mathbf{h}^{\text{meta}}_{(s_0,r,o_0,t_0)}$ as follows:
\begin{equation}
\label{eq: relproj}
\resizebox{0.88\columnwidth}{!}{%
$
    \mathbf{h}_{r} = f_{\text{proj}}\left(\mathbf{h}^{\text{meta}}_{(s_0,r,o_0,t_0)} + f_{r}^2\left(\sigma \left(f_{r}^1\left(\mathbf{h}^{\text{meta}}_{(s_0,r,o_0,t_0)}\right)\right)\right)\right),
$
}
\end{equation}
where $f_{r}^1:\mathbb{R}^{2d} \to \mathbb{R}^{4d}$, $f_{r}^2:\mathbb{R}^{4d} \to \mathbb{R}^{2d}$, $f_{\text{proj}}:\mathbb{R}^{2d} \to \mathbb{R}^{\frac{d}{2}}$ are three single layer neural networks. The meta-representation $\mathbf{h}_{r} \in \mathbb{R}^{\frac{d}{2}}$ will then be used in the metric function to compute the scores of the TKG quadruples. 

\subsubsection{Metric Function}
To compute the plausibility score for the query quadruple $(s_q,r,o_q,t_q)$, 
our metric function requires the time-aware entity representations $\mathbf{h}_{(s_q,t_q)}$, $\mathbf{h}_{(o_q,t_q)}$ of the query entities $s_q$, $o_q$ at $t_q$, as well as the meta-representation of $r$. We derive $\mathbf{h}_{(s_q,t_q)}$, $\mathbf{h}_{(o_q,t_q)}$ as follows:
\begin{equation}
\label{eq: queryemb}
\resizebox{0.6\columnwidth}{!}{%
$
\begin{aligned}
    &\mathbf{h}_{(s_q,t_q)} = \mathbf{h}_{s_q}+ \delta_2 f(\mathbf{h}_{s_q}\|\boldsymbol{\Phi}(t_q)),\\
    &\mathbf{h}_{(o_q,t_q)} = \mathbf{h}_{o_q}+ \delta_2 f(\mathbf{h}_{o_q}\|\boldsymbol{\Phi}(t_q)).
\end{aligned}
$
}
\end{equation}
$\mathbf{h}_{s_q} \in \mathbb{R}^d$ and $\mathbf{h}_{o_q} \in \mathbb{R}^d$ denote the time-invariant entity representations of $s_q$ and $o_q$, respectively.
$\delta_2$ is a trainable parameter controlling the amount of the injected temporal information. Different from Equation \ref{eq: agg}, we do not search temporal neighbors from the background graph for query entities and no aggregation is performed.
Equation \ref{eq: queryemb} states how we compute query entity representations in MOST-TA. Similarly, MOST-TD adapts Equation \ref{eq: queryemb} to the following form to enable time difference learning, i.e., $\mathbf{h}_{(s_q,t_q)} = \mathbf{h}_{s_q}+ \delta_2 f(\mathbf{h}_{s_q}\|\boldsymbol{\Phi}(t_q-t_0))$; $\mathbf{h}_{(o_q,t_q)} = \mathbf{h}_{o_q}+ \delta_2 f(\mathbf{h}_{o_q}\|\boldsymbol{\Phi}(t_q-t_0))$.

Inspired by the KG scoring function RotatE \cite{sunrotate}, we design a metric function that computes scores based on complex vectors, and treat the meta-representation of $r$ as element-wise rotation in the complex plane. 
To achieve this, we first transform $\mathbf{h}_{r}$ into a complex vector $\mathbf{h}_r^{\mathbb{C}} \in \mathbb{C}^{\frac{d}{2}}$. 
\begin{equation}
\label{eq: queryrel}
\resizebox{0.88\columnwidth}{!}{%
$
\begin{aligned}
    &\Tilde{\mathbf{h}}_r = \frac{\pi}{\lVert\mathbf{h}_r\rVert_\infty} \mathbf{h}_r,\\
    &\mathbf{h}_r^{\mathbb{C}}[j] = \cos\left(\Tilde{\mathbf{h}}_r[j]\right) + \sqrt{-1}\sin\left(\Tilde{\mathbf{h}}_r[j]\right), \quad 1<=j<=\frac{d}{2}.
\end{aligned}
$
}
\end{equation}
$\mathbf{h}_r^{\mathbb{C}}[j]$ and $\Tilde{\mathbf{h}}_r[j]$ denote the $j$th element of the vectors $\mathbf{h}_r^{\mathbb{C}}$ and $\Tilde{\mathbf{h}}_r$, respectively, e.g., $\mathbf{h}_r^{\mathbb{C}} = \left[ \mathbf{h}_r^{\mathbb{C}}[1], \dots, \mathbf{h}_r^{\mathbb{C}}[\frac{d}{2}] \right]^\top$. $\lVert\mathbf{h}_r\rVert_\infty$ denotes the infinity norm of the vector $\mathbf{h}_r$. 
We then map the query entity representations $\mathbf{h}_{(s_q,t_q)}$, $\mathbf{h}_{(o_q,t_q)}$ to the complex space $\mathbb{C}^{\frac{d}{2}}$ and get $\mathbf{h}_{(s_q,t_q)}^{\mathbb{C}}$, $\mathbf{h}_{(o_q,t_q)}^{\mathbb{C}}$.
The real part of each mapped vector is the first half of the original vector from $\mathbb{R}^d$ and the imaginary part is the second half. For example, if $\mathbf{h}_{(s_q,t_q)} = [2,3]^\top \in \mathbb{R}^2$, we map it to $\mathbf{h}_{(s_q,t_q)}^{\mathbb{C}} = [2+3\sqrt{-1}]^\top \in \mathbb{C}^1$. Since unitary complex number can be regarded as a rotation in the complex plane (explained in \cite{sunrotate}), $\mathbf{h}_{(s_q,t_q)}^{\mathbb{C}} \circ \mathbf{h}_r^{\mathbb{C}}$ can be interpreted as doing element-wise rotation from the query subject $s_q$ in the complex plane. 
Based on it, the complete form of our metric function $\psi$ is given as 
\begin{equation}
\label{eq: metric score}
\resizebox{0.85\columnwidth}{!}{%
$
    \psi\left(q | \mathcal{S}_r \right) = \text{Sigmoid}\left(\textbf{Re}\left( \left(\mathbf{h}_{(s_q,t_q)}^{\mathbb{C}} \circ \mathbf{h}_r^{\mathbb{C}} \right)^\top \Bar{\mathbf{h}}_{(o_q,t_q)}^{\mathbb{C}} \right)\right),
$
}
\end{equation}
where $q = (s_q,r,o_q,t_q)$ and $\text{Sigmoid}$ denotes the sigmoid function. $\textbf{Re}$ means taking the real part of the complex number and $\Bar{\mathbf{h}}_{(o_q,t_q)}^{\mathbb{C}}$ means the complex conjugate of $\mathbf{h}_{(o_q,t_q)}^{\mathbb{C}}$. $\psi$ takes the real part of the dot product (Hermitian product) between $\mathbf{h}_{(s_q,t_q)}^{\mathbb{C}} \circ \mathbf{h}_r^{\mathbb{C}}$ and $\mathbf{h}_{(o_q,t_q)}^{\mathbb{C}}$ as the plausibility score of $q$ given $\mathcal{S}_r$. Note that $\psi$ can be viewed as doing similarity matching of support and query quadruples (the higher the score is, the more $q$ resembles the support quadruple). 
Thus, MOST is taken as a metric-based meta-learning approach.

\subsubsection{Parameter Learning}
We train MOST with episodic training. In each episode, we sample one sparse relation $r$.
Then we sample an $r$-related quadruple as its support quadruple $\mathcal{S}_r = (s_0,r,o_0,t_0)$, 
and collect a group of query quadruples containing $r$. For each of the query quadruple, i.e., $q = (s_q,r,o_q,t_q) \in \mathcal{Q}_r$, we further switch $q$'s object entity $o_q$ to every other entity $e \in (\mathcal{E}\setminus\{o_q\})$ in the TKG ($\mathcal{E}$ denotes the set of all entities in this TKG) and construct $|\mathcal{E}|-1$ polluted quadruples $\{q^-\}$ for $q$. Then we use the binary cross entropy loss to optimize our model.
\begin{equation}
\resizebox{0.7\columnwidth}{!}{%
$
    \mathcal{L} = \frac{1}{|\mathcal{Q}_r|}\sum_{q} \frac{1}{|\mathcal{E}|} \left( l_{q} + \sum_{q^-} l_{q^-}\right),
$
}
\end{equation}
where $l_{q} =  -y_{q} \log \left(\psi(q|\mathcal{S}_r)\right) - (1-y_{q})\log (1 -$ $\psi(q|\mathcal{S}_r) )$ and $l_{q^-} = -y_{q^-} \log \left(\psi(q^-|\mathcal{S}_r)\right) - (1-y_{q^-}) \log \left(1-\psi(q^-|\mathcal{S}_r) \right)$ denote the binary cross entropy loss of $q$ and $q^-$, respectively. $y_{q} = 1$ and $y_{q^-} = 0$ because for $q \in \mathcal{Q}_r$, we want its score $\psi(q|\mathcal{S}_r)$ to be maximized, while $q^-$ is a polluted quadruple, and thus we want its score $\psi(q^-|\mathcal{S}_r)$ to be minimized.
We describe our one-shot training procedure with Algorithm \ref{alg: ont shot}.
\begin{algorithm}[t]
\scriptsize
\caption{One-Shot Episodic Training}
\label{alg: ont shot}

\DontPrintSemicolon
\KwInput{Training sparse relations $\mathcal{R}_{sp}^{train}$}

\For{\textup{episode = 1: M}}
{
Shuffle relations in $\mathcal{R}_{sp}^{train}$

Sample sparse relation $r$ from $\mathcal{R}_{sp}^{train}$

Sample a $(s_0,r,o_0,t_0)$ to make the support set $\mathcal{S}_r$

\If {\textup{One-Shot Interpolated LP}}
{Sample a batch of query quadruples $\mathcal{Q}_r = \{(s_q,r,o_q,t_q)\}$}

\Else(\tcp*[h]{One-Shot Extrapolated LP})
{Sample a batch of query quadruples $\mathcal{Q}_r = \{(s_q,r,o_q,t_q) | t_0 < t_q\}$}

Compute $\mathbf{h}_{(s_0,t_0)}$, $\mathbf{h}_{(o_0,t_0)}$ with graph encoder

Compute meta-information $\mathbf{h}^{\text{meta}}_{(s_0,r,o_0,t_0)}$

Learn meta-representation $\mathbf{h}_{r}$ with meta-representation learner

Pollute each $q \in \mathcal{Q}_r$ and generate polluted quadruples $\{q^-\}$ 

Compute time-aware representations for entities in all $q$ and $\{q^-\}$ \tcp*[h]{Equation \ref{eq: queryemb}}

Compute scores for all $q$ and $\{q^-\}$ with metric function $\psi$

Calculate the loss $\mathcal{L}$

Update model parameters using gradient of loss $\bigtriangledown	\mathcal{L}$
}

\end{algorithm}




\section{Experiments}
\label{sec: experiments}
We evaluate MOST and several baselines on our newly-proposed datasets (Section \ref{sec:exp results}).
We analyze model components with ablation studies in Section \ref{sec: model analysis}. We also compare MOST with several strong baselines over the performance of different sparse relations and different support-query time difference $|t_q-t_0|$ in Section \ref{appendix: diff rel} and \ref{appendix: diff sup que diff}, respectively.
\begin{table*}[t]
\caption{Experimental results of both one-shot TKG interpolated and extrapolated LP on all four newly-proposed datasets. Evaluation metrics are filtered MRR and Hits@1/5/10. The best results are marked in bold.}
\label{tab: tid int results}
    \centering
    \resizebox{\textwidth}{!}{
    \small\begin{tabular}{@{}cccccccccccccccccc@{}}
\toprule
        \multicolumn{9}{c}{\textbf{One-Shot TKG Interpolated Link Prediction}} & \multicolumn{9}{c}{\textbf{One-Shot TKG Extrapolated Link Prediction}}\\
\cmidrule(lr){1-9} \cmidrule(lr){10-18} 
        \textbf{Datasets} & \multicolumn{4}{c}{\textbf{ICEWS-one\_int}} & \multicolumn{4}{c}{\textbf{GDELT-one\_int}} &
        \textbf{Datasets} & \multicolumn{4}{c}{\textbf{ICEWS-one\_ext}} & \multicolumn{4}{c}{\textbf{GDELT-one\_ext}}\\
\cmidrule(lr){2-5} \cmidrule(lr){6-9}  \cmidrule(lr){11-14} \cmidrule(lr){15-18}
        \textbf{Model} & MRR & Hits@1 & Hits@5 & Hits@10 & MRR & Hits@1 & Hits@5 & Hits@10 &
        \textbf{Model} & MRR & Hits@1 & Hits@5 & Hits@10 & MRR & Hits@1 & Hits@5 & Hits@10\\
\midrule 
        TNTComplEx & 23.34 & 14.57 & 31.54 & 36.88 
        & 11.95 & 6.76  & 15.58 & 21.78
        & TANGO & 10.23 & 3.94 & 15.88 & 25.78 
        & 13.88 & 9.61 & 16.93 & 22.29
         \\
        ATiSE & 34.40 & 22.03  & 49.25 & 60.57 
        & 7.77 & 5.10  & 8.13 & 12.13
        & CyGNet & 22.30 & 12.61  & 30.46 & 39.13 
        & 9.42 & 4.87  & 13.13 & 16.81
         \\
        TeLM  & 35.38 & 24.42  & 47.74 & 59.12 
        & 10.41 & 5.97  & 13.28 & 18.87
        & xERTE  & 30.02 & 19.79  & 42.13 & 51.16 
        & 16.38 & 10.88 & 22.19 & 27.76
         \\
        
\midrule
        GANA  & 13.83 & 6.07  & 24.00 & 27.23  
       & 5.89 & 2.53   & 8.35 & 12.20
       & GANA  & 11.34 & 3.70 & 19.25 & 25.67  
       & 7.12 & 4.85 & 8.89 & 11.13
       \\
       MetaR  & 27.69 & 7.88  & 52.58 & 61.78  
       & 9.91 & 0.18  & 19.62 & 26.79
       & MetaR  & 23.50 & 9.01  & 40.18 & 48.73  
       & 9.66 & 0.03   & 19.52 & 26.30
       \\
         GMatching & 30.59 & 15.46  & 48.80 & 58.62  
         & 12.53 & 6.55  & 17.14 & 24.15
         & GMatching & 20.30 & 12.35  & 28.80 & 38.02 
         & 12.26 & 8.41  & 13.76 & 19.01
         \\
    FSRL & 33.98 & 18.94  & 52.61 & 59.82
       & 14.11 & 7.61  & 19.56 & 27.54
       & FSRL & 18.06 & 12.09  & 21.06 & 32.23
       & 6.96 & 2.52  & 11.58 & 14.13
       \\
       
       FAAN & 35.48 & 23.27  & 49.45 & 57.73  
       & 14.77 & 7.67  & 21.35 & 27.11
       & FAAN & 25.73 & 15.86  & 35.95 & 43.73  
       & 14.36 & 8.71  & 18.46 & 23.71
        \\
\midrule     
      OAT  & 32.45 & 24.26  & 39.88 & 48.43
      & 12.26 & 7.58  & 15.53 & 21.46
      & OAT  & 16.25 & 7.44  & 25.38 & 33.16
      & 7.47 & 2.96  & 7.01 & 12.59
        \\       
\midrule
        MOST-TA 
        & \textbf{47.79} & \textbf{39.91}  & \textbf{57.01} & 62.25
        & \textbf{17.71} & 11.56 & \textbf{23.25} & \textbf{29.76}
        & MOST-TA
        & 32.94 & 26.35  & 39.97 & 47.19
        & 15.69 & 10.14  & 20.54 & 26.38
        \\ 
        & $\pm$ 0.1 & $\pm$ 0.2  & $\pm$ 0.3 & $\pm$ 0.3
        & $\pm$ 0.1 & $\pm$ 0.1 & $\pm$ 0.1 & $\pm$ 0.2 
        &
        & $\pm$ 0.2 & $\pm$ 0.3  & $\pm$ 0.3 & $\pm$ 0.5
        & $\pm$ 0.1 & $\pm$ 0.1 & $\pm$ 0.1 & $\pm$ 0.1 
        \\
        MOST-TD 
        & 47.60 & 39.43  & 56.83 & \textbf{62.38}
        & 17.36 & \textbf{11.67}  & 22.74 & 28.63
        & MOST-TD 
        & \textbf{38.46} & \textbf{31.51}  & \textbf{46.02} & \textbf{52.32}
        & \textbf{17.36} & \textbf{11.64}  & \textbf{22.46} & \textbf{28.15}
        \\
        & $\pm$ 0.2 & $\pm$ 0.2  & $\pm$ 0.4 & $\pm$ 0.3
        & $\pm$ 0.1 & $\pm$ 0.1 & $\pm$ 0.1 & $\pm$ 0.1 
        &
        & $\pm$ 0.2 & $\pm$ 0.3  & $\pm$ 0.4 & $\pm$ 0.5
        & $\pm$ 0.1 & $\pm$ 0.1 & $\pm$ 0.1 & $\pm$ 0.1 
        \\
\bottomrule
    \end{tabular}
    }
\end{table*}
\subsection{Experimental Setup}
\subsubsection{Evaluation Metrics} We employ two evaluation metrics, i.e., Hits@1/5/10 and mean reciprocal rank (MRR), to evaluate model performance. For each query quadruple $(s_q, r, o_q, t_q) \in \mathcal{Q}_r$, $r \in \mathcal{R}_{sp}^{test}$, we derive an object prediction query: $(s_q, r, ?, t_q)$. We compute the rank of the ground truth missing entity $o_q$ for every object prediction query based on the scores computed with score function $\psi$. Let $\text{rank}_{o_q}$ denote the rank of $o_q$ in $(s_q, r, ?, t_q)$. We compute MRR by averaging the reciprocal of ranks among all the query quadruples in the meta-test set: $\frac{1}{\sum_{r \in \mathcal{R}_{sp}^{test}} |\mathcal{Q}_{r}|}\sum _{r \in \mathcal{R}_{sp}^{test}} \sum_{q \in \mathcal{Q}_{r}}  \frac{1}{\text{rank}_{o_q}}$,
where $q$ denotes a query quadruple $(s_q,r,o_q,t_q)$ in the meta-test set. Note that if $r \in \mathcal{R}_{sp}^{test}$, then its reciprocal relation $r^{-1} \in \mathcal{R}_{sp}^{test}$. Performing object prediction over the query quadruples related to $r^{-1}$ equals performing subject prediction over the query quadruples related to $r$. The restriction to only considering object prediction in MRR computation will not lead to a loss of generality. Hits@1/5/10 are the proportions of the predicted links where ground truth entities are ranked as top 1, top 5, top 10, respectively. We follow \cite{bordes2013translating} and use filtered results for fairer evaluation. 

\subsubsection{Baseline Methods and Implementation Details}
We consider five static KG FSL methods, i.e., Gmatching \cite{xiong2018one}, MetaR \cite{chen2019meta}, FSRL \cite{zhang2020few}, FAAN \cite{sheng2020adaptive}, GANA \cite{niu2021relational}, and one TKG FSL method, i.e., OAT \cite{mirtaheri2021oneshot}. 
We provide static KG FSL methods with all the facts in the original datasets, and neglect time information, i.e., neglecting $t$ in $(s,r,o,t)$. 
Besides, three traditional TKG interpolation methods, i.e., TNTComplEx \cite{lacroixtensor}, ATiSE \cite{xu2020temporal}, TeLM \cite{xu2021temporal}, and three traditional TKG extrapolation methods, i.e., TANGO \cite{han2021learning}, CyGNet \cite{zhu2021learning}, xERTE \cite{han2021explainable} are considered. For each interpolation dataset, we build a training set for these methods by adding all the background quadruples in $\mathcal{G}'$ and the quadruples regarding every $r \in \mathcal{R}_{sp}^{train}$. We further add the support quadruple associated with each sparse relation $r \in (\mathcal{R}_{sp}^{valid} \cup \mathcal{R}_{sp}^{test})$ into the training set. For each extrapolation dataset, we build a training set by adding all the background quadruples during meta-training time and the quadruples concerning every $r \in \mathcal{R}_{sp}^{train}$. 
We do not include any quadruple regarding $r \in (\mathcal{R}_{sp}^{valid} \cup \mathcal{R}_{sp}^{test})$ into the training set due to the time constraint in the extrapolation setting, 
but we allow the models to use the support quadruples ($\mathcal{S}_r$, $r \in (\mathcal{R}_{sp}^{valid} \cup \mathcal{R}_{sp}^{test})$) during inference. We test all methods over same test quadruples to ensure fair comparison. We implement all baselines with their official open-sourced implementations. We do all experiments (including baselines and MOST) with PyTorch \cite{paszke2019pytorch} on a single NVIDIA Tesla T4. All experimental results are obtained with the mean of 5 runs with different random seeds.
\subsection{Experimental Results}
\label{sec:exp results}
Table \ref{tab: tid int results} reports the experimental results of one-shot TKG interpolated and extrapolated LP. 
MOST outperforms baseline methods on all datasets in both LP tasks. 
Traditional TKG reasoning methods are not FSL methods so it is hard for them to model sparse relations given only one associated quadruple. 
Static KG FSL methods do not consider temporal information so they are weaker than MOST. When computing support entity representations, the TKG FSL method OAT includes temporal information by employing a snapshot encoder that sequentially encodes a small number of historical graph snapshots right before the support timestamp $t_0$. In interpolated LP, OAT loses information coming after $t_0$, and in extrapolated LP, a short history length fails to include the abundant information outside the considered history. MOST searches for a fixed number of temporal neighbors of support entities in its graph encoder and impose no constraint on how temporally far away these neighbors are. This helps to incorporate temporally farther information. Besides, OAT employs cosine similarity for score computation, which is beaten by our metric function. Table \ref{tab: tid int results} also shows that while MOST-TA beats MOST-TD in interpolated LP, MOST-TD outperforms MOST-TA in extrapolated LP. For MOST-TA, in the interpolated LP, part of the TKG at every timestamp is observable during training, thus enabling the time encoder to learn from all timestamps. However, in the extrapolated LP, meta-training set does not span across the whole timeline, leading to MOST-TA's degenerated performance during inference when we sample the temporal neighbors from the timestamps unseen in the meta-training set. For extrapolation, modeling time differences (MOST-TD) achieves better results since almost all time differences we encounter during inference are already seen and learned during meta-training.
\begin{table}[htbp]
    \caption{Ablation studies of MOST variants on ICEWS-based datasets. The best results are marked in bold. 
    }\label{tab: ablation}
    \centering
    \resizebox{\columnwidth}{!}{
    \small\begin{tabular}{@{}lcccccccccccc@{}}
\toprule
        & \multicolumn{6}{c}{\textbf{MOST-TA}} & \multicolumn{6}{c}{\textbf{MOST-TD}}\\
\cmidrule(lr){2-7} \cmidrule(lr){8-13}
        \textbf{Datasets} & \multicolumn{3}{c}{\textbf{ICEWS-one\_int}} &  \multicolumn{3}{c}{\textbf{ICEWS-one\_ext}}
        & \multicolumn{3}{c}{\textbf{ICEWS-one\_int}} &  \multicolumn{3}{c}{\textbf{ICEWS-one\_ext}}\\
\cmidrule(lr){2-4} \cmidrule(lr){5-7} \cmidrule(lr){8-10} \cmidrule(lr){11-13}
        \textbf{Variants} & MRR & Hits@1 & Hits@10 & MRR & Hits@1 & Hits@10
        & MRR & Hits@1 & Hits@10 & MRR & Hits@1 & Hits@10\\
\midrule 
        A  & 9.21 & 3.05  & 21.95 & 9.87 & 3.04 & 26.69 
         & 9.21 & 3.05  & 21.95 & 9.87 & 3.04 & 26.69
        \\
\midrule 
        B1  & 45.00 & 35.40  & 61.83 & 29.98 & 20.17 & 46.30
        & 44.91 & 34.56  & 60.40 & 35.45 & 27.87 & 51.77
         \\
         B2  & 42.99 & 32.14  & 61.47 & 31.78 & 24.40  & 46.19
         & 42.11 & 30.45 & 61.79 & 37.73 & 30.62 & 52.32 
         \\
\midrule 
        C1  & 1.00 & 0.82  & 0.89 & 9.33 & 5.16 & 17.38 
         & 0.95 & 0.62  & 0.97 & 10.46 & 6.17 & 15.77
         \\
         C2  & 16.27 & 7.36  & 32.47 & 26.13 & 17.97 & 43.65
         & 23.68 & 13.87  & 44.40 & 27.35 & 18.82 & 43.78
         \\
         
\midrule
        MOST&  \textbf{47.79} & \textbf{39.91} & \textbf{62.25} & \textbf{32.94}
        & \textbf{26.35} & \textbf{47.19}
        &  \textbf{47.60} & \textbf{39.43} & \textbf{62.38} & \textbf{38.46}
        & \textbf{31.51} & \textbf{52.32}
        \\
\bottomrule
    \end{tabular}
    }

\end{table}
\begin{table*}[htbp]
    \caption{One-shot TKG interpolated LP performance over each sparse relation on ICEWS-one\_int and GDELT-one\_int. The best results are marked in bold. The second best results are underlined.}\label{tab: diff rel int}
    \centering
    \resizebox{\textwidth}{!}{
    \small\begin{tabular}{@{}cccccccccc@{}}
\toprule
        \multicolumn{5}{c}{\textbf{ICEWS-one\_int}} & \multicolumn{5}{c}{\textbf{GDELT-one\_int}}\\
\cmidrule(lr){1-5}\cmidrule(lr){6-10}
         & & \multicolumn{3}{c}{\textbf{MRR}}  & & & \multicolumn{3}{c}{\textbf{MRR}} \\
\cmidrule(lr){3-5}\cmidrule(lr){8-10}
        \textbf{Relation} & \textbf{Frequency} & MOST-TA & FAAN & TeLM & \textbf{Relation} & \textbf{Frequency} & MOST-TA & FAAN & TeLM \\
\midrule
        Threaten to reduce or break relations & 65 & \textbf{35.25} & \underline{28.13} & 21.79 &
        Receive deployment of peacekeepers & 108 & \textbf{35.94} & 9.52 & \underline{10.62}\\
        
        Demonstrate for leadership change & 89 &  \textbf{53.65} & 40.86 & \underline{44.08} &
        Ban political parties or politicians & 167 &  \underline{9.46} & \textbf{11.54} & 5.30\\
        
        Express intent to yield & 92 & \textbf{50.50} & \underline{36.38} & 26.06 &
        Attempt to assassinate & 177 & \textbf{12.23} & \underline{7.65} & 6.36\\
        
        Increase police alert status & 118 & \underline{75.75} & 53.06 & \textbf{78.31}&
        Receive inspectors & 234 & \textbf{17.07} & \underline{15.13} & 1.26\\
        
        Appeal for material aid & 146 & \textbf{52.69} & \underline{49.37} & 37.96 &
        Demand change in institutions, regime & 411 & \textbf{14.16} & 7.00 & \underline{15.42}\\
        
        Impose blockade, restrict movement & 175 & \textbf{63.83} & 49.41 & \underline{56.58} &
        Threaten political dissent & 675 & \textbf{18.21} & 13.77 & \underline{16.15}\\
        
        Impose restrictions on political freedoms & 282 & \textbf{38.45} & \underline{34.33} & 23.00 &
        Declare truce, ceasefire & 748 & \textbf{19.65} & \underline{13.41} & 6.33\\
        
        Acknowledge or claim responsibility & 269 & \textbf{39.54} & 25.66 & \underline{38.48} &
        Give ultimatum & 752 & \textbf{17.98} & \underline{16.84} & 11.47\\
        
        Share intelligence or information & 349 & \textbf{29.75} & 15.18 & \underline{27.21} &
        &&&\\
        Defy norms, law & 436 & \textbf{57.84} & \underline{43.37} & 29.33 &
        &&&\\
\bottomrule
    \end{tabular}
    }

\end{table*}
\begin{table*}[htbp]
    \caption{One-shot TKG extrapolated LP performance over each sparse relation on ICEWS-one\_ext and GDELT-one\_ext. The best results are marked in bold. The second best results are underlined.}\label{tab: diff rel ext}
    \centering
    \resizebox{\textwidth}{!}{
    \small\begin{tabular}{@{}cccccccccc@{}}
\toprule
        \multicolumn{5}{c}{\textbf{ICEWS-one\_ext}} & \multicolumn{5}{c}{\textbf{GDELT-one\_ext}}\\
\cmidrule(lr){1-5}\cmidrule(lr){6-10}
         & & \multicolumn{3}{c}{\textbf{MRR}}  & & & \multicolumn{3}{c}{\textbf{MRR}} \\
\cmidrule(lr){3-5}\cmidrule(lr){8-10}
        \textbf{Relation} & \textbf{Frequency} & MOST-TD & FAAN & xERTE & \textbf{Relation} & \textbf{Frequency} & MOST-TD & FAAN & xERTE \\
\midrule
        Provide military protection or peacekeeping & 55 & \textbf{32.85} & \underline{17.63} & 17.10 &
        Investigate crime, corruption & 149 & \underline{15.34} & \textbf{15.90} & 15.29 \\
        
        Accuse of human rights abuses & 57 & \textbf{21.39} & 4.99 & \underline{17.07} &
        Express intent to de-escalate military engagement & 153 & \textbf{19.58} & 15.83 & \underline{18.22} \\
        
        Appeal for change in leadership & 65 & \underline{22.05} & 21.60 & \textbf{30.98} &
        Express intent to settle dispute & 160 & \textbf{22.19} & 8.01 & \underline{21.58} \\
        
        Acknowledge or claim responsibility & 67 & \textbf{26.31} & 23.30 & \underline{25.23} &
        Protest violently, riot & 176 & \textbf{18.72} & 11.66 & \underline{13.42} \\
        
        Share intelligence or information & 93 & \textbf{30.41} & 18.52 & \underline{26.35} &
        Carry out suicide bombing & 180 & \textbf{10.17} & 5.06 & \underline{10.01} \\
        
        Rally opposition against & 107 & \underline{21.67} & 15.40 & \textbf{24.33} &
        Seize or damage property & 212 & \underline{17.04} & 11.43 & \textbf{17.29} \\
        
        Express intent to provide material aid & 127 & \textbf{37.13} & 28.26 & \underline{29.09} &
        Veto & 279 & \underline{22.78} & \textbf{23.15} & 22.67 \\
        
        Appeal for intelligence cooperation & 128 & \underline{42.91} & \textbf{43.88} & 41.67 &
        Demand intelligence cooperation & 312 & 14.75 & \textbf{15.61} & \underline{15.31} \\
        
        Provide military aid & 130 & \textbf{33.51} & 9.74 & \underline{18.59} &
        Engage in political dissent & 321 & \textbf{15.67} & 10.21 & \underline{11.83} \\
        
        Mobilize or increase armed forces & 180 & \textbf{59.61} & \underline{32.96} & 28.99 &
        Appeal for economic aid & 348 & 15.21 & \textbf{16.17} & \underline{15.55} \\
        
        Bring lawsuit against & 184 & \textbf{47.47} & 36.28 & \underline{45.99} &
        Express intent to provide economic aid & 359 & \textbf{19.69} & 17.67 & \underline{18.72} \\
\bottomrule
    \end{tabular}
    }

\end{table*}
\subsection{Ablation Studies}
\label{sec: model analysis}
To validate the effectiveness of model components, we conduct several ablation studies with MOST on ICEWS-based datasets. We present the experimental results in Table \ref{tab: ablation}. Studies are designed from the following angles:
\textbf{(A) Excluding temporal information:} In A, we remove the time encoder $\boldsymbol{\Phi}$ in all model components (Equation \ref{eq: agg}, \ref{eq: agg2} and \ref{eq: queryemb}), creating a model without using any temporal information. Note that without $\boldsymbol{\Phi}$, MOST-TA equals MOST-TD. We observe that it is crucial to utilize temporal information in MOST for achieving strong results in one-shot TKG LP.
\textbf{(B) Changing graph aggregation function in meta-information extractor:} 
In B1, we employ mean pooling over time-aware representations of temporal neighbors, e.g., Equation \ref{eq: agg} becomes $\mathbf{h}_{(s_0,t_0)} = \frac{1}{|\Tilde{\mathcal{N}}_{s_0}|} \sum_{(e',r', t') \in \Tilde{\mathcal{N}}_{s_0}}f(\mathbf{h}_{e'}\|\boldsymbol{\Phi}(t'))$ for $s_0$. In B2, we employ RGCN \cite{schlichtkrull2018modeling} coupled with $\boldsymbol{\Phi}$, e.g., Equation \ref{eq: agg} becomes $\mathbf{h}_{(s_0,t_0)} = \frac{1}{|\Tilde{\mathcal{N}}_{s_0}|} \sum_{(e',r', t') \in \Tilde{\mathcal{N}}_{s_0}}\mathbf{W}_{r'}\left(f(\mathbf{h}_{e'}\|\boldsymbol{\Phi}(t'))\right)$ for $s_0$. We observe that our graph aggregation function helps to effectively capture meta-information. \textbf{(C) Changing metric function:} In C1, we switch our metric function $\psi$ to RotatE, i.e., Equation \ref{eq: metric score} becomes $\|\mathbf{h}_{(s_q,t_q)}^{\mathbb{C}} \circ \mathbf{h}_r^{\mathbb{C}} - \mathbf{h}_{(o_q,t_q)}^{\mathbb{C}}\|_1$, where $\|\cdot\|_1$ is the L1-norm. Note that we input time-aware entity representations into RotatE and thus C1 also achieves temporal reasoning. In C2, we switch $\psi$ to the LSTM-based matcher proposed in \cite{xiong2018one}. We perform two steps of matching. Each step of matching is defined as $\mathbf{h}'_{k+1}, \mathbf{c}_{k+1} = \text{LSTM}(\mathbf{h}_{\text{query}},[\mathbf{h}_{k}\|\mathbf{h}_{\text{support}},\mathbf{c}_{k}]);
 \mathbf{h}_{k+1} = \mathbf{h}'_{k+1} + \mathbf{h}_{\text{query}}; \text{score}_{k+1} = \mathbf{h}_{k+1}^\top \mathbf{h}_{\text{support}}$.
$\text{LSTM}(\mathbf{x},[\mathbf{h}, \mathbf{c}])$ is a standard LSTM cell \cite{DBLP:journals/neco/HochreiterS97} with input $\mathbf{x}$, hidden state $\mathbf{h}$ and cell state $\mathbf{c}$. $\mathbf{h}_{\text{support}} = \mathbf{h}_{(s_0,t_0)}\|\mathbf{h}_{(o_0,t_0)}$, and $\mathbf{h}_{\text{query}} = \mathbf{h}_{(s_q,t_q)}\|\mathbf{h}_{(o_q,t_q)}$ are time-aware for temporal reasoning. We find that our metric function works much better in our tasks. 
\subsection{Performance over Different Sparse Relations}
\label{appendix: diff rel}

We compare MOST with several strong baselines regarding the performance over different sparse relations. We choose to compare with the strongest traditional TKG interpolation and extrapolation baselines, i.e., TeLM and xERTE. We further choose to compare with the strongest KG FSL baseline, i.e., FAAN, since it outperforms almost all baselines in our main results in Table \ref{tab: tid int results}. 
We aggregate the object prediction results of each original relation and its reciprocal relation to get the overall subject and object prediction results of the original relation. For example, for the original sparse relation $r$, we aggregate the object prediction results of $r$ and $r^{-1}$ to get the overall subject and object prediction results of $r$. We report the overall results in Table \ref{tab: diff rel int} and Table \ref{tab: diff rel ext}.
From Table \ref{tab: diff rel int}, we observe that MOST outperforms FAAN and TeLM in almost all relations, showing its strong robustness over different sparse relations in one-shot TKG interpolated LP. From Table \ref{tab: diff rel ext}, we observe that MOST also shows strong robustness in one-shot TKG extrapolated LP by outperforming FAAN and xERTE in most sparse relations.

\subsection{Performance over Different Support-Query Time Differences}
\label{appendix: diff sup que diff}
\begin{figure*} 
    \centering
  \subfloat[\label{fig: most_faan-ICEWS05-15-INT}]{%
       \includegraphics[width=0.25\textwidth]{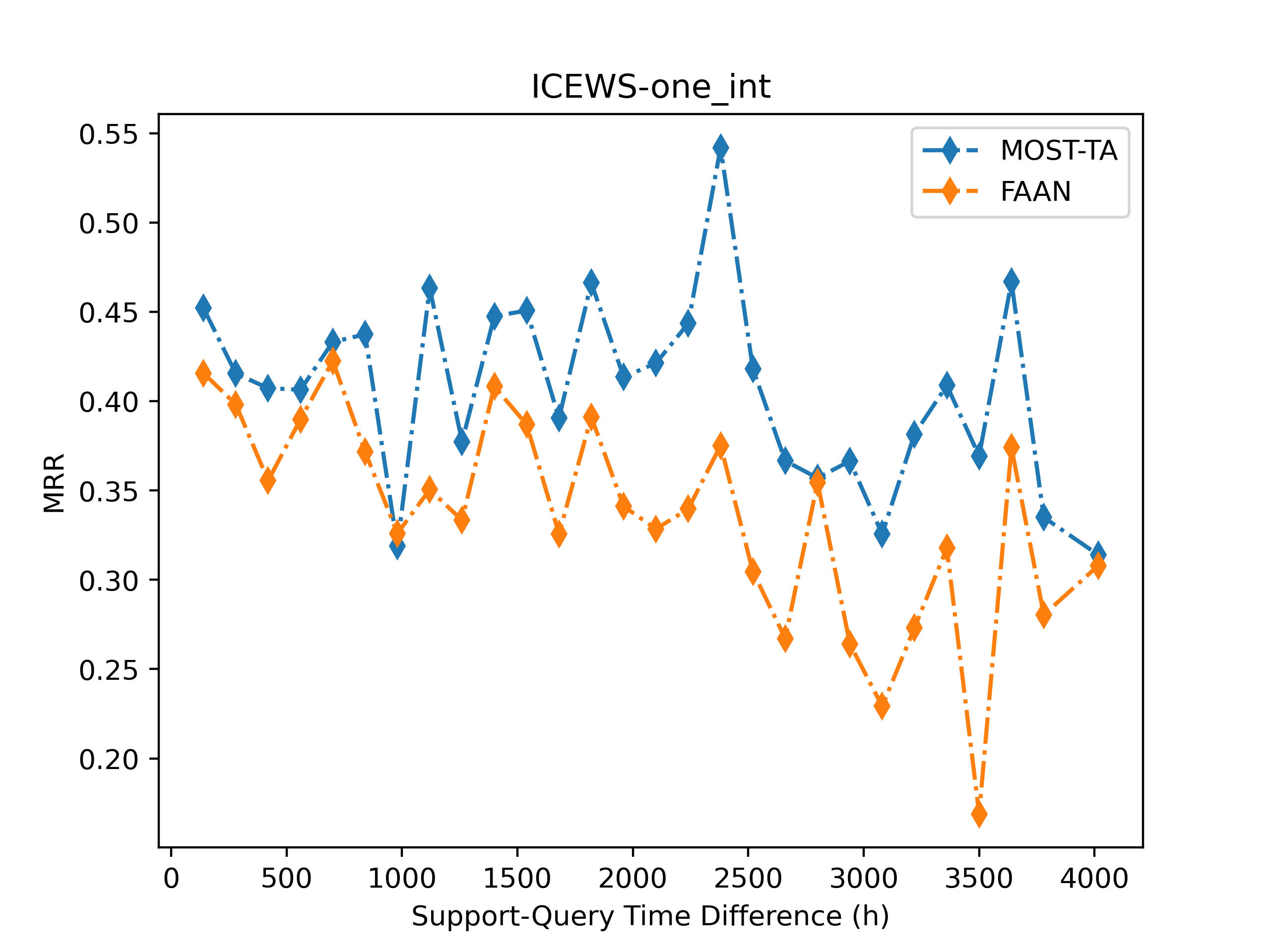}}
    \hfill
  \subfloat[\label{fig: most_telm-ICEWS05-15-INT}]{%
        \includegraphics[width=0.25\textwidth]{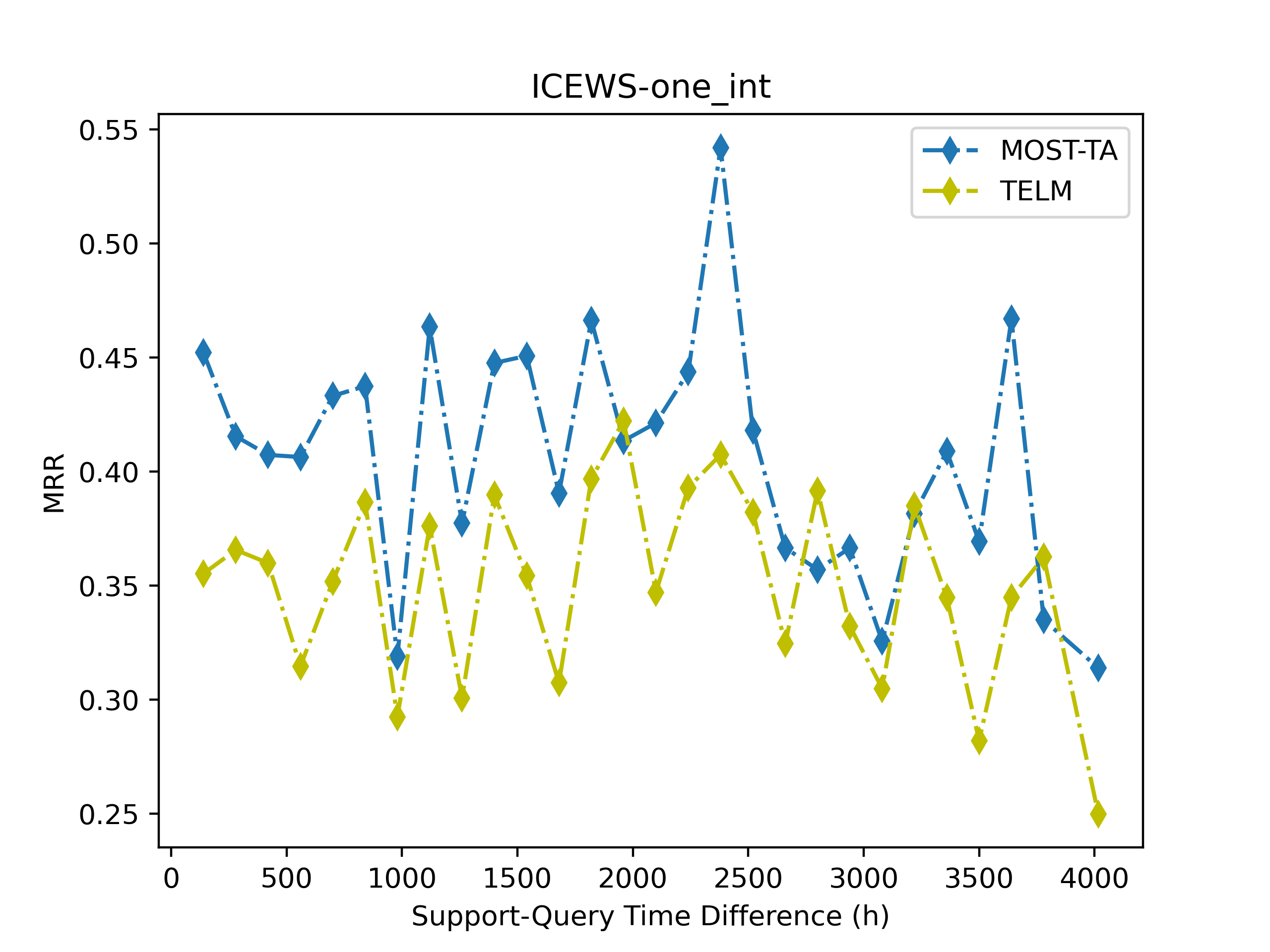}}
  \subfloat[\label{fig: most_faan-GDELT-INT}]{%
        \includegraphics[width=0.25\textwidth]{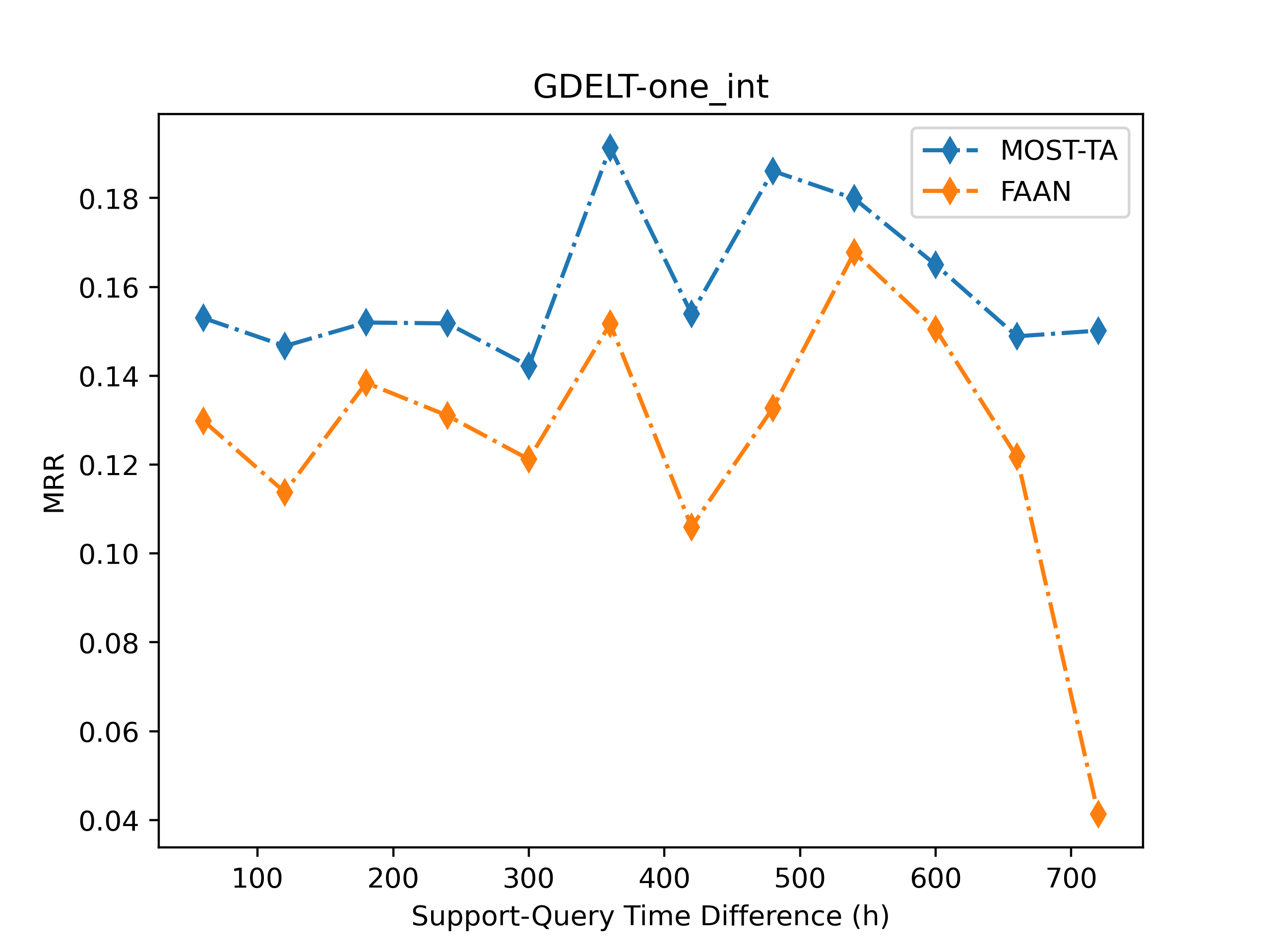}}
    \hfill
  \subfloat[\label{fig: most_telm-GDELT-INT}]{%
        \includegraphics[width=0.25\textwidth]{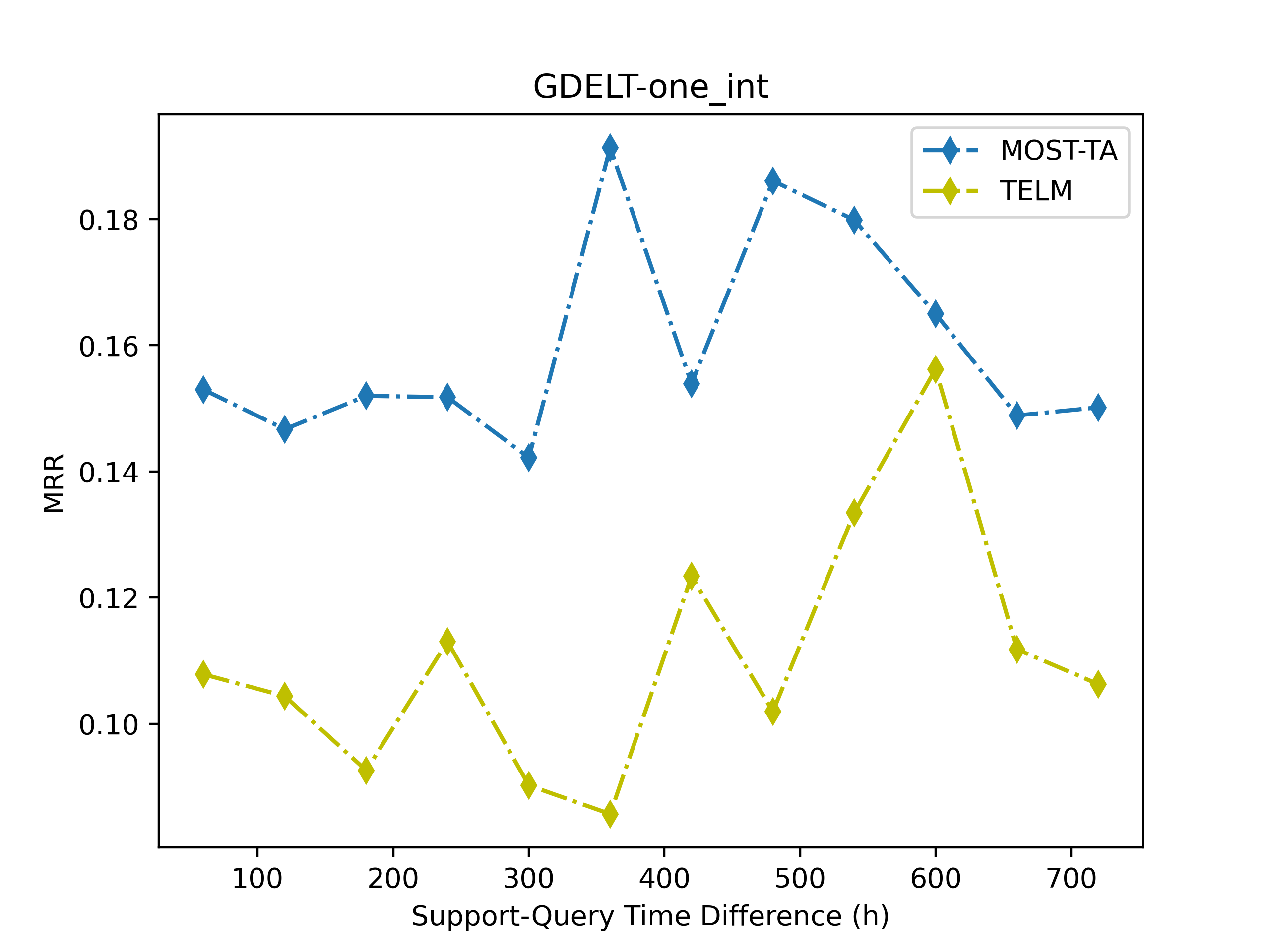}}
    \hfill
  \subfloat[\label{fig: most_faan-ICEWS05-15-EXT}]{%
        \includegraphics[width=0.25\textwidth]{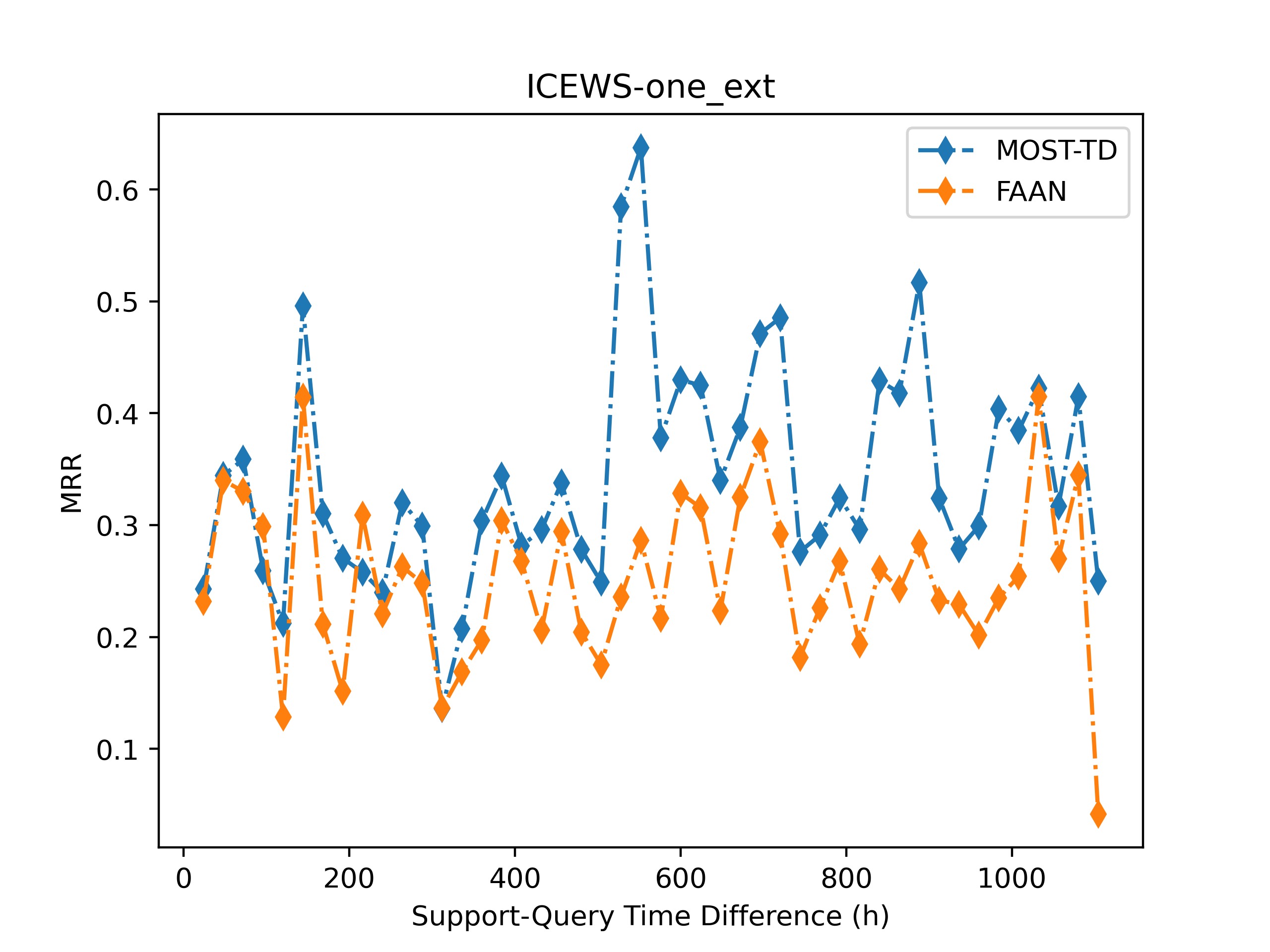}}
    \hfill
  \subfloat[\label{fig: most_xerte-ICEWS05-15-EXT}]{%
        \includegraphics[width=0.25\textwidth]{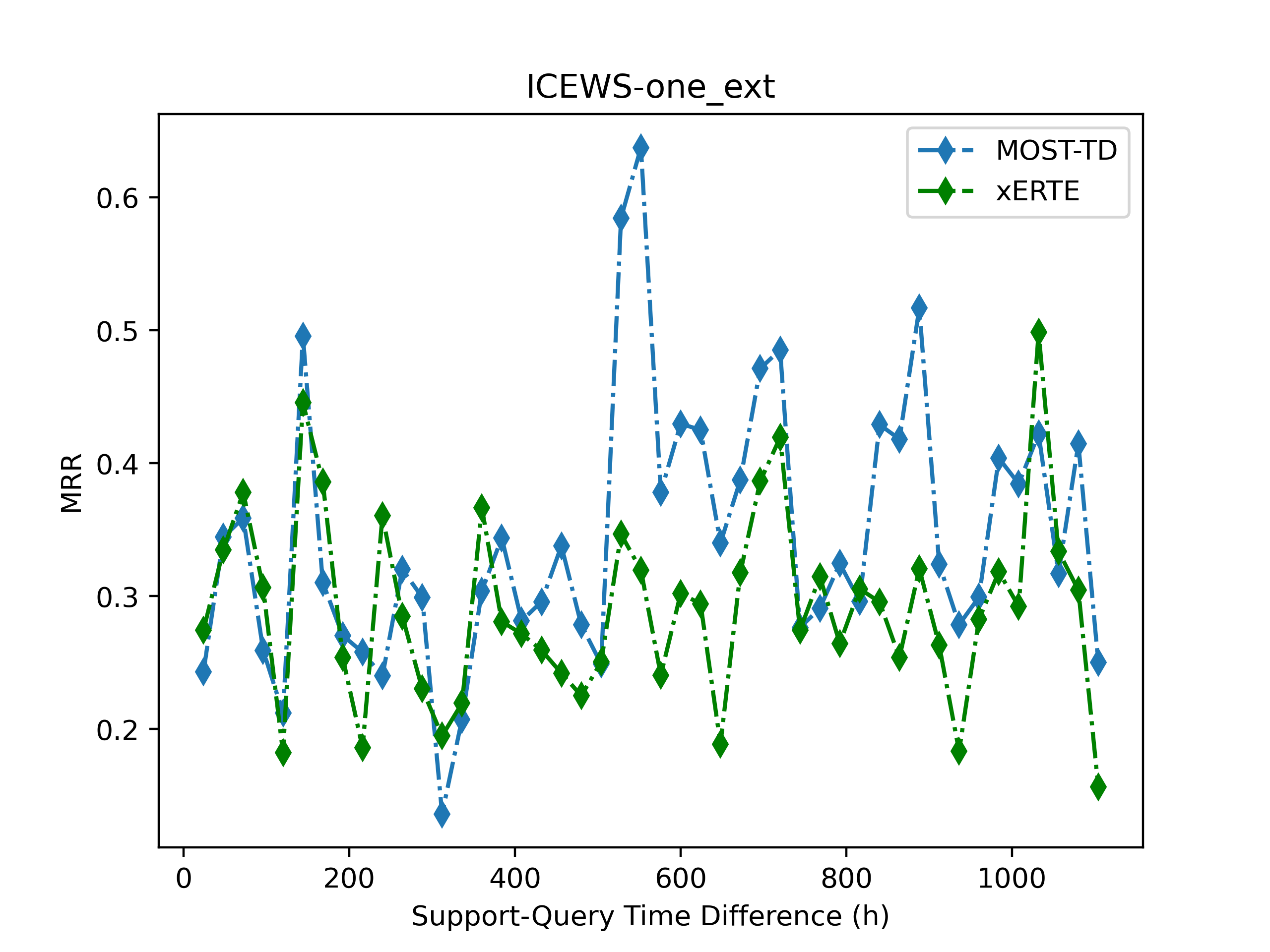}}
    \hfill
  \subfloat[\label{fig: most_faan-GDELT-EXT}]{%
        \includegraphics[width=0.25\textwidth]{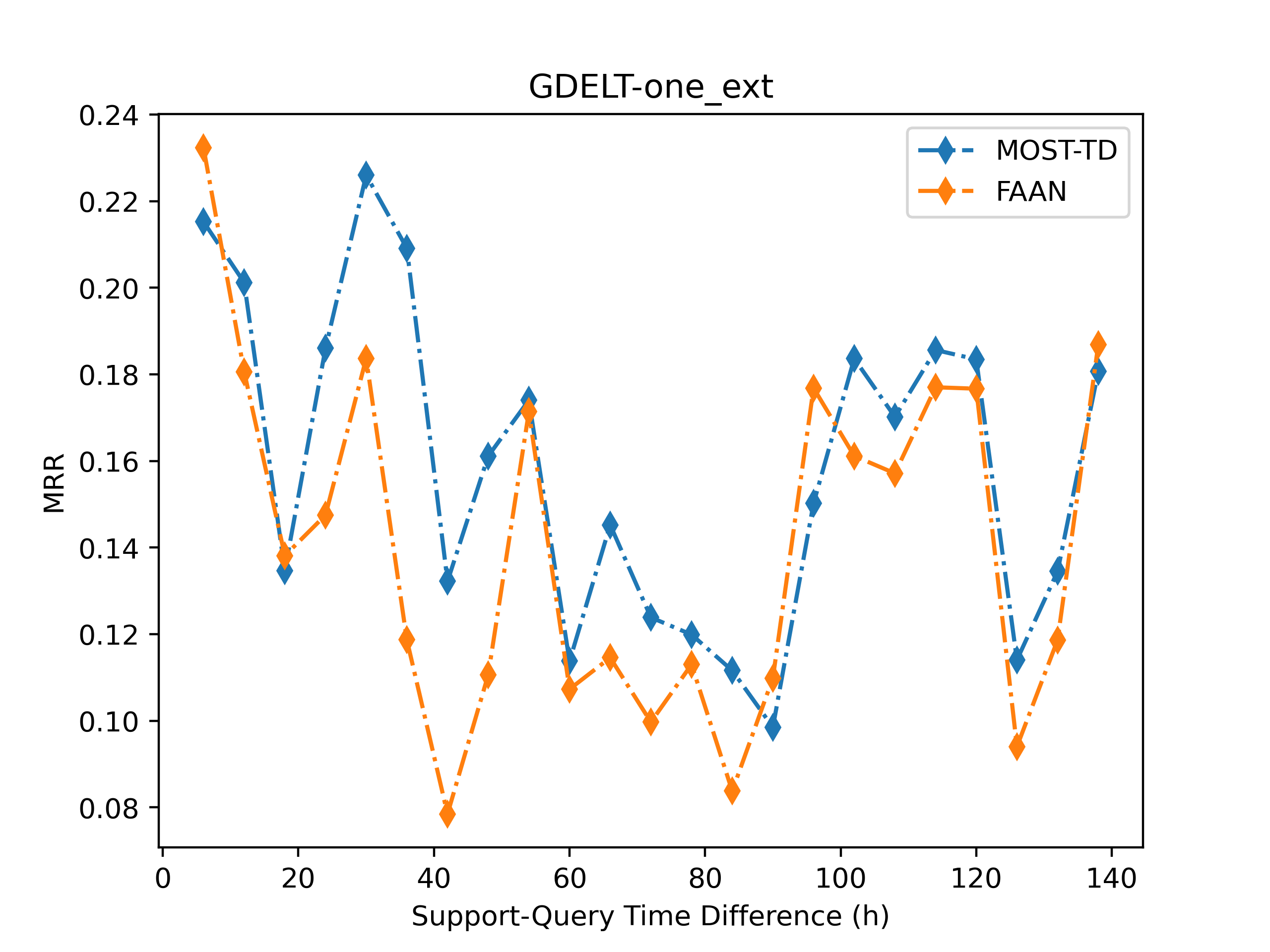}}
    \hfill
  \subfloat[\label{fig: most_xerte-GDELT-EXT}]{%
       \includegraphics[width=0.25\textwidth]{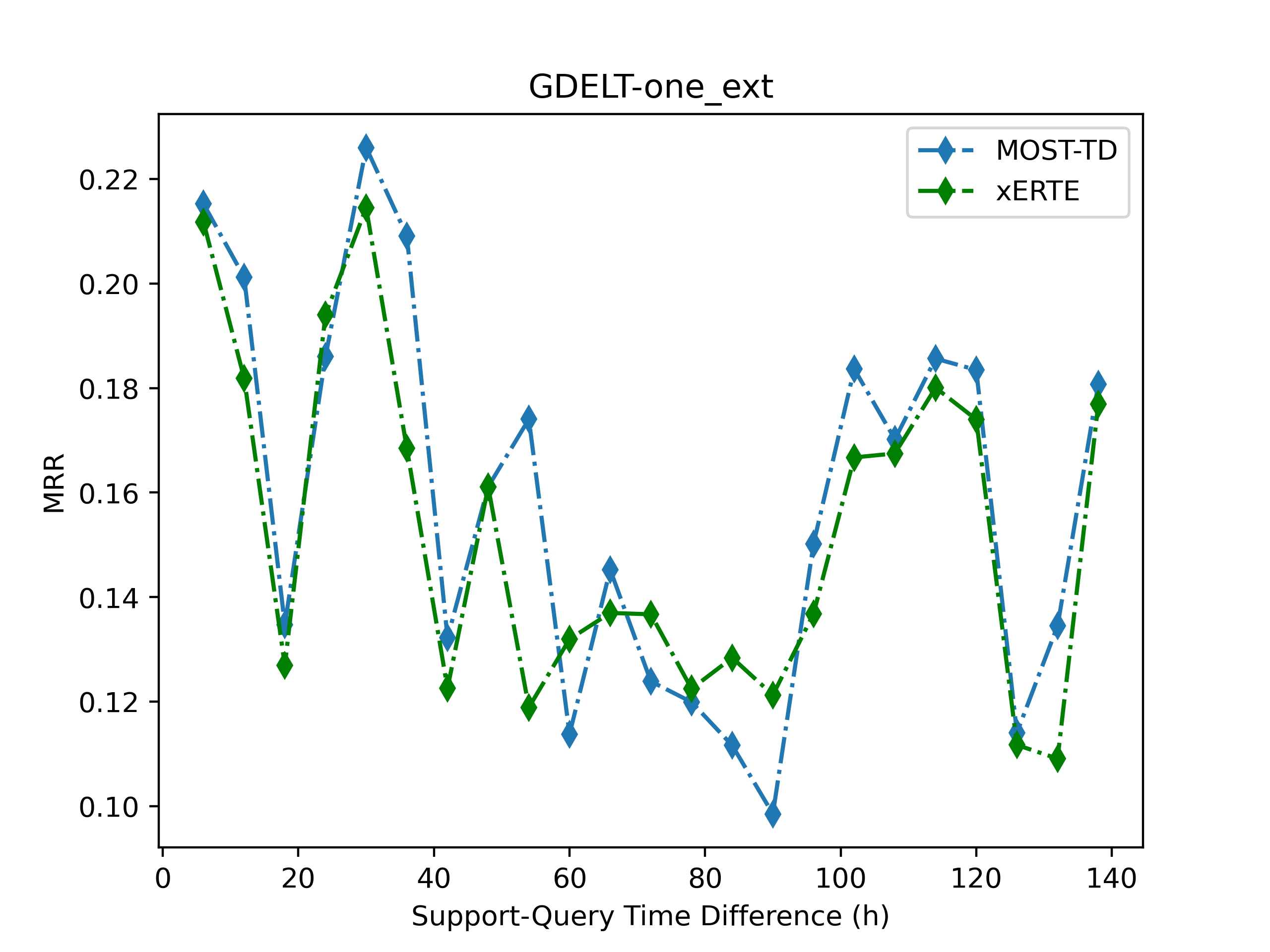}}
    \hfill
  \caption{Performance comparison between MOST and baselines over different support-query time differences $|t_q-t_0|$. (a) MOST-TA vs. FAAN on ICEWS-one\_int; (b) MOST-TA vs. TeLM on ICEWS-one\_int; (c) MOST-TA vs. FAAN on GDELT-one\_int; (d) MOST-TA vs. TeLM on GDELT-one\_int; (e) MOST-TD vs. FAAN on ICEWS-one\_ext; (f) MOST-TD vs. xERTE on ICEWS-one\_ext; (g) MOST-TD vs. FAAN on GDELT-one\_ext; (h) MOST-TD vs. xERTE on GDELT-one\_ext.}
  \label{fig: sup que diff} 
\end{figure*}
We compare MOST with several strong baselines regarding the performance over different support-query time differences. We choose to compare with TeLM, xERTE and FAAN due to the reasons explained in Section \ref{appendix: diff rel}. For each sparse relation $r$ in the meta-test set, we compute the time difference between the support timestamp $t_0$ and every query timestamp $t_q$. We want to study whether our model is robust to different $|t_q - t_0|$ by plotting the performance of MOST and the selected baseline methods according to the test quadruples with different values of $|t_q - t_0|$ (Fig. \ref{fig: sup que diff}). Since the test quadruples are all associated to sparse relations, the number of test quadruples who share the same value of $|t_q - t_0|$ is prone to be small.
Thus, following \cite{mirtaheri2021oneshot}, we aggregate every 140 hours for ICEWS-one\_int by combining each model's performance over the test quadruples whose $|t_q - t_0|$ lie within each time interval (140 hours) to form a point in the plot representing each model's overall test performance over them. For example, the first orange point from the left in Fig. \ref{fig: most_faan-ICEWS05-15-INT} represents the overall performance of FAAN over the test quadruples whose $|t_q - t_0|$ lie within 0 and 140 hours. Similarly, we aggregate every 24 hours for ICEWS-one\_ext. For GDELT-based datasets, we aggregate every 40 hours on GDELT-one\_int and every 6 hours on GDELT-one\_ext. 
We ensure that the original model performance is reported to plot Fig. \ref{fig: sup que diff} and every model's performance is aggregated in the same way. From Fig. \ref{fig: most_faan-ICEWS05-15-INT} to \ref{fig: most_telm-GDELT-INT}, we observe that MOST constantly outperforms FAAN and TeLM on both one-shot TKG interpolated LP datasets. From Fig. \ref{fig: most_faan-ICEWS05-15-EXT} to \ref{fig: most_xerte-GDELT-EXT}, we find that MOST outperforms FAAN and xERTE in most cases on both one-shot TKG extrapolated LP datasets. To this end, we show that MOST is robust to different support-query time differences $|t_q - t_0|$, indicating the effectiveness of its temporal reasoning components.

\section{Conclusion}
We extend both TKG interpolated and extrapolated LP tasks to the one-shot scenario, fix the unreasonable task setting employed by previous work, and propose a model learning \textbf{m}eta-representations of \textbf{o}ne-\textbf{s}ho\textbf{t} relations for solving both tasks (MOST). MOST learns sparse relations' meta-representations based on the time-aware representations of the entities in the one-shot examples. It further employs a metric function for predicting missing entities from the unobserved TKG facts regarding sparse relations. To overcome the problem brought by previous datasets, we further propose four large-scale datasets for one-shot TKG LP. We compare MOST with recent baselines on our new datasets. Experimental results show that MOST achieves superior performance. MOST is designed only for one-shot learning and cannot be directly used in multi-shot cases, e.g., 3-shot. It is worthwhile to draw attention to the multi-shot scenario of TKG LP over sparse relations in the future. 
\bibliography{custom}
\bibliographystyle{IEEEtran}

\end{document}